%% file: ICIP_SS_PINN_AMD.tex
\documentclass{article}
\usepackage{spconf,amsmath,graphicx}

\input{macros_gpi}

\def\rfbnn{\rfb_{\tiny NN}}
\def\bvi{\blue{v_{i}}}
\def\rwb{\red{\wb}}
\def\rwbh{\red{\hat{\wb}}}
\def\rgbh{\red{\gbh}}

\def\Red#1{{#1}}
\def\red#1{{#1}}
\def\blue#1{{#1}}

\title{BAYESIAN PINNs FOR UNCERTAINTY-AWARE INVERSE PROBLEMS (BPINN-IP)}
\name{Ali MOHAMMAD-DJAFARI}
\address{ISCT, Bures-sur-Yvette, France \\ 
Institute of Digital Twin (IDT), EIT, Ningbo, China \\
Dept. of Statistics, Central South University, Changcha, China}

\begin{document}
%
\maketitle
\begin{abstract}
The main contribution of this paper is to develop a hierarchical Bayesian formulation of PINNs for linear inverse problems, which is called BPINN-IP. The proposed methodology extends PINN to account for prior knowledge on the nature of the expected NN output, as well as its weights. Also, as we can have access to the posterior probability distributions, naturally uncertainties can be quantified. 
Also, variational inference and Monte Carlo dropout are employed to provide predictive means and variances for reconstructed images. Un example of applications to deconvolution and super-resolution is considered, details of the different steps of implementations are given, and some preliminary results are presented. 
\end{abstract}
\section{Introduction}
\label{sec:intro}

Inverse problems are fundamental to numerous scientific and engineering disciplines, where the objective is to infer hidden variables or parameters from indirect and often noisy observations. Applications span a broad spectrum, including medical and biological imaging \cite{Jin2017, Ker2018}, geophysical exploration \cite{Waheed2021, Puzyrev2019}, and industrial non-destructive testing \cite{Miya2002}.

Classically, inverse problems have been handled using analytical methods, regularization techniques, and Bayesian inference 
\cite{engl1996regularization,Tanner2019}. Bayesian inference, in particular, offers a way to handle uncertainties by modeling the posterior distribution of unknowns given observed data \cite{Jiang2017}. However, these methods can become computationally prohibitive, especially in high-dimensional settings or when the forward model involves complex physics.

The advent of neural networks (NN) and deep learning (DL) has introduced new possibilities for solving inverse problems, with neural networks capable of approximating complex mappings between observations and unknown parameters \cite{Raissi2019, Yang2021}. Physics-Informed Neural Networks (PINNs) extend this concept by embedding physical laws as constraints within the learning process, improving generalization and reducing the need for large training datasets \cite{Raissi2019, Haghighat2021}.

This paper extends the PINN framework by integrating Bayesian inference, leading to the Bayesian Physics-Informed Neural Network (BPINN). This approach not only recovers point estimates of unknown parameters but also provides uncertainty quantification via posterior distributions.

In this paper, we build upon these advancements, developing a Bayesian PINN framework tailored for linear inverse problems. By combining the interpretability and physical consistency of PINNs with the uncertainty quantification of Bayesian inference, we aim to deliver a practical, robust approach for both simulated and experimental scenarios. 

\section{Proposed Hierarchical Bayesian PINN framework}
\label{sec:proposed-model}

The proposed methodology can be summarized in the following five steps: 

\smallskip\noindent{1. }
First step is to assume that we have a forward or generative model:
\beq
\bgb=\Hb(\rfb)+\epsilonb
\eeq
where $\rfb$ is the input, $\bgb$ the output, $\Hb$ the forward model, and $\epsilonb$ the errors, both measurement and modeling. 

\smallskip\noindent{2. }
Second step is to assign priors $p(\rfb)$ and $p(\epsilonb)$. Using this second one, and the forward model, we can deduce the likelihood 
$p(\bgb|\rfb)$. Then, we can use the Bayes rule to obtain the posterior law:  
\beq
p(\rfb|\bgb)=\frac{p(\bgb|\rfb) p(\rfb)}{p(\bgb)}
\eeq
For the case of linear inverse problem: $\bgb=\Hb\rfb+\epsilonb$, Gaussian likelihooh $p(\bgb|\rfb)=\Nc(\bgb|\Hb\rfb,\bve\Ib)$, and Gaussian prior 
$p(\rfb)=\Nc(\rfb|\bar{\fb},\bvf\Ib)$, the posterior $p(\rfb|\bgb)$ is also Gaussian: 
\beq
p(\rfb|\bgb)=\Nc(\rfb|\rfbh,\rSigmabh)
\eeq
with
\beq
\left\{\barr{l}
\rfbh=[\Hb'\Hb+\lambda\Ib]^{-1}\Hb'(\bgb-\Hb\bar{\fb})\\\
\rSigmabh=\bve[\Hb'\Hb+\lambda\Ib]^{-1}, \quad \lambda=\frac{\bve}{\bvf}
\earr\right. 
\eeq

\smallskip\noindent{3.}
Third step is to design an appropriate neural network which, takes as input $\bgb$ and as output $\rfbnn$. This NN can be considered as a proxy, surrogate or approximate inversion, such that its output $\rfbnn$ can be very near to the true or ground truth $\rfb$. 

\smallskip\noindent{4. }
Fourth step is training the NN. We distinguish two cases: 
\bit
\item Supervised case where we have a set of outputs-inputs $\{\bgb_{Ti}, \bfb_{Ti}\}$, and 
\item Unsupervised case where the available data are only $\{\bgb_{Ti}\}$. 
\eit
The parameters of the NN, $\rwb$ are estimated at the end of training step. 

\smallskip\noindent{5. }
Fifth step is using the trained NN, giving it any other input $\bgb_j$ which produces the outputs $\rfbh_j$ which we hope to be not very far from the ground truth $\rfb_j$. We may also want to be able to quantify the its uncertainty. 

To go more in details, first we consider the supervised case, and then the unsupervised case. 

\subsection{Supervized training data}

First we consider the supervised case, i.e., when we have a set of labeled  training data, (output-input of the generating forward system) : 
$\{\bgb_{Ti}, \bfb_{Ti}\}$. 
\begin{align*}
\btabu{@{}c@{}} Data \\ $\{\bgb_{Ti}, \bfb_{Ti}\}$\etabu
\Ra\fbox{~NN ($\rwb$)~}\Ra\rfb_{NNi}\Ra\fbox{~$\Hb$~}\Ra\rgbh_{NNi} 
\end{align*}
To write, step by step, the relations and equations, we follow: 

\smallskip\noindent{1. }
We assume that, given the true $\rfb$ and the forward model 
$\Hb$, the the data $\{\bgb_{Ti}\}$ are generated independently:
\beq
p(\{\bgb_{Ti}\}|\{\rfb_i\})=\prod_i p(\bgb_{Ti}|\rfb_i),
\eeq 
and we can assign the likelihood $p(\bgb_{Ti}|\rfb_i)$, for example, the Gaussian:  
\beq
p(\bgb_{Ti}|\rfb_i,\bve_i\Ib) = \Nc(\bgb_{Ti}|\Hb\rfb_i,\bve_i\Ib).
\eeq
The same for $\{\bfb_{Ti}\}$, we assume that, given the true $\rfb$, we have:
\beq
p(\{\bfb_{Ti}\}|\{\rfb_i\})=\prod_i p(\bfb_{Ti}|\rfb_i)=\prod_i \Nc(\bfb_{Ti}|\rfb_i,\bvf_i\Ib)
\eeq 

\smallskip\noindent{2. }
We also assign a prior $p(\rfb)$ to the true $\rfb$ to translate our prior knowledge about it, and using the Bayes rule we get: 
\begin{align}
p(\rfb_i|\{\bgb_{Ti}, \bfb_{Ti}\}) \propto \prod_i & p(\bgb_{Ti}|\rfb_i,\bvf_i\Ib) \nonumber \\
& p(\bfb_{Ti}|\rfb_i,\bvf_i\Ib) \, p(\rfb_i) 
\end{align}

\smallskip\noindent{3. }
With the Gaussian cases of the equations [5, 6, 7], we can rewrite the posterior as: 
\begin{align}
p(\rfb_i|\{\bgb_{Ti},\bfb_{Ti}\}) &\propto \expf{-J(\rfb_i)} \nonumber \\ 
\mbox{with:~~} 
J(\rfb_i)= \sum_i & \frac{1}{2\bve_i}\|\bgb_{Ti}-\Hb\rfb_i\|^2 + \nonumber \\
& \frac{1}{2\bvf_i}\|\bfb_{Ti}-\rfb_i\|^2 + \ln p(\rfb_i)
\end{align}

Choosing also a Gaussian prior: $p(\rfb_i)=\Nc(\rfb_i|\bar{\bfb},\bvi\Ib)$, we get: 
\begin{align}
J(\rfb_i)=
\sum_i & 
\frac{1}{2\bvf_i} \|\bfb_{Ti}-\rfb_i\|^2 + \nonumber \\
& \frac{1}{2\bve_i}\|\bgb_{Ti}-\Hb\rfb_i\|^2 + 
\frac{1}{2\bvi} \|\rfb_i-\bar{\bfb}\|^2
\end{align}
It is then easy to see that the $p(\rfb_i|\{\bgb_{Ti},\bfb_{Ti}\})$ is Gaussian: 
\beq
p(\rfb_i|\{\bgb_{Ti},\bfb_{Ti}\})=\Nc(\rfb_i|\rfbh_i,\rSigmabh_i)
\eeq
with
\beq
\left\{\barr{l}
\rfbh_i=\sum_i [\Hb'\Hb+\lambda_i\Ib]^{-1}\Hb'(\bgb_{Ti}-\Hb\bfb_{Ti}-\Hb\bar{\rfb}) \\[3pt]
\rSigmabh_i=\sum_i\bve_i[\Hb'\Hb+\lambda_i\Ib]^{-1}, \quad 
\lambda_i={\bve_i}/{\bvf_i}
\earr\right.
\eeq
\smallskip\noindent{4. }
Now, we consider the training step of the NN where we note the output of the neural network $\rfb_{NNi}$ which is a function of NN parameters $\rwb$ and the input data $\{\bgb_{Ti}\}$, we can define the following optimization criterion: 
\begin{align}
& J(\rwb)=
\sum_i \frac{1}{\bvf_i} \|\bfb_{Ti}-\rfb_{NNi}(\rwb)\|^2 + \nonumber \\ 
& \frac{1}{\bve_i}\|\bgb_{Ti}-\Hb\rfb_{NNi}(\rwb)\|^2 + \frac{1}{\bve_i}\|\bar{\bfb} - \rfb_{NNi}(\rwb)\|^2 
\end{align}
This criterion can be considered as a physics based or physics informed neural network (PINN), 
where the classical output residual part is: 
\beq
J_{NN}(\rwb)= \sum_i\frac{1}{2\bve_i} \|\bfb_{Ti}-\rfb_{NNi}(\rwb)\|^2
\eeq 
and the physics informed part is:
\begin{align}
J_{PI}(\rwb)=\sum_i & \frac{1}{2\bve_i} \|\bgb_{Ti}-\Hb\rfb_{NNi}(\rwb)\|^2+ \nonumber \\ 
& \frac{1}{2\bvi}\|\bar{\bfb} - \rfb_{NNi}(\rwb)\|^2 
\end{align}

\smallskip\noindent{5. }
We can also consider 
\beq
p(\rwb|\{\bgb_{Ti},\bfb_{Ti}\}) \propto \expf{-J(\rwb) - \ln p(\rwb)}
\eeq
and use it as the posterior probability distribution of the NN's parameter $\rwb$ given the training data $\{\bgb_{Ti},\bfb_{Ti}\}$. A specific choice for the prior $p(\rwb)$ can be one of the sparsity enforcing priors, such as 
\beq
p(\rwb) \propto \expf{-\gamma_w \|\rwb\|_\beta^\beta}.
\eeq
However, in practice, the full expression of this posterior is very difficult to obtain, and even its optimization can be more difficult if the NN contains nonlinear activation function. 

\subsection{Unsupervised training data}
In the unsupervised case, we only have a set of training data: $\{\bgb_{Ti}\}$. The classical NN methods can not be applied as there is not any reference data. The main advantage of the PINN is exactly in the fact that, it can be applied in this case. The schematic for this case is almost the same, but we have only for training data $\{\bgb_{Ti}\}$: 
\begin{align*}
\btabu{@{}c@{}} Data \\ $\{\bgb_{Ti}\}$\etabu
\Ra\fbox{~NN ($\rwb$)~}\Ra\rfb_{NNi}\Ra\fbox{~$\Hb$~}\Ra\rgbh_{NNi}  
\end{align*}
However, we can still follow the same steps we had for the supervised case, and obtain: 
\begin{align}
&p(\rwb|\{\bgb_{Ti}\}) \propto \expf{-J(\rwb)} \nonumber\\ 
\mbox{with:~} 
&J(\rwb)=\sum_i \frac{1}{\bve_i}\|\bgb_{Ti}-\Hb\rfb_i(\rwb)\|^2 + 
\gamma_w\|\rwb\|_1
\end{align}
which can be used for the estimation of the NN's parameters. 

In both supervised and unsupervised cases, we still have to choose an appropriate structure for the NN, its depth, its number of hidden variable, as well as choosing appropriate optimization algorithms, it learning rate, etc. These difficulties make the implementation of such methods not very easy and many experience are needed to implement, to train and to use them in real applications. 

\subsection{Inference and uncertainty quantification}
Once the NN has been trained, we use it to infer $\rfb$ for new data $\bgb_j$:
\[
\bgb_j \;\Ra\; \fbox{\btabu{c} Inference step \\ NN ($\rwbh$) \etabu} \;\Ra\; \rfbh_{NNj}~~.
\]
The hope is that $\rfbh_{NNj}$ is close to the (unknown) ground truth $\rfb_j$, for example in terms of
\[
\Delta = \|\rfbh_{NNj} - \rfb_j\|^2.
\]
Beyond a point estimate, we would like to quantify the uncertainty of $\rfbh_{NNj}$. The inference step with uncertainty quantification (UQ) can be summarized as
\[
\bgb_j \;\Ra\; \fbox{\btabu{c} Inference step \\ with UQ \\ NN ($\rwbh$)\etabu}
\;\Ra\;
\left\{\barr{c} \rfbh_{NNj} \\[6pt] \Sigmabh_{NNj} \earr\right.,
\]
where $\rfbh_{NNj} = \mathbb{E}[\rfb_j|\bgb_j]$ is the posterior mean, and $\Sigmabh_{NNj}$ its covariance. As $\rfb \in \mathbb{R}^N$ is high dimensional, $\Sigmabh_{NNj} \in \mathbb{R}^{N\times N}$ is even larger, and in practice we often restrict attention to its diagonal (pixelwise variances). For imaging problems, one can thus visualize both a mean image and a variance (or standard deviation) image; examples will be provided in the Results section.

\section{Application in infrared image processing}
\label{sec:application}

We considered two applications: infrared image restoration and super-resolution. These two problems are two classical inverse problems. 
In the first one, the forward model can be written as:
\begin{equation}
\bg(x,y)=h(x,y)*\phi(\rf(x,y)) +\epsilon(x,y),
\end{equation}
where $g(x,y)$ is the observed infrared image, $h(x,y)$ is the point spread function (psf) of the imaging system (diffusion of heat from source to camera and the camera response itself), $f(x,y)$ is the unknown temperature distribution, and $\epsilon(x,y)$ represents the errors. 
When discretized, this relation can be wriiten as:
\begin{equation}
\bgb=\Hb\Phib\rfb +\epsilonb,
\end{equation}
where $\bgb$ contains all the pixels of the image, $\Hb$ is the 2D convolution matrix, $\rfb$ contains all the pixels of the temperature distribution, $\epsilonb$ the errors, and $\Phib$ is a diagonal matrix of the nonlinear point heat transfert equation. 

In super-resolution, the forward model can also be summerised as: 
\begin{equation}
\bgb=\Hb\Db\rfb+\epsilonb,
\end{equation}
where, here $\rfb$ is the high resolution (HR) image, $\Db$ is the down sampling operation, $\Hb$ a filtering operation, and $\bgb$ the observed low resolution (LR) image.  

As we can see, the implementation of these two problems are almost the same. To implement the BPINN for these inverse problems, we followed: 
\begin{itemize}
\item Generate a set of synthetic images to create the training and testing database. 
In Figure~1, we see one of these data (Original and Blurred) images for the image restoration, and in Figure~2, an example of image superresolution.

\item Construct an appropriate NN structure. 
In these simulations, we used a simple NN with three hidden layers and RELU activiation functions.

\item Define the loss functions, as explained in the proposed method, the optimisation algorithm, here a gradient based, all the necessary parameters such as learning rate, etc, and train the model.

\item Measure the performances of the trained model during the training, validation and testing.

\item Save the optimized trained model for ulterior use.

\item Load it and use it for real input images.
\end{itemize}

A Python Jupyter notebook implementing all steps is available and will be publicly released upon acceptance of the paper.

Here, we only show a few results of the implemented methods on one simulated and on real image. 
Figure~1 shows an example of IR image restoration and Figure~2 an example of superresolution. In both cases, we generated 1000 synthetic images of size 128x128 for training data set, 800 of them are used for training and 200 for testing. The NN for both cases is a CNN with few number of layes. The loss functions are, in general has two or three parts, as explained in details in the proposed model section of the paper. 

\begin{figure}[h!]
\begin{center}
\includegraphics[scale=.35]{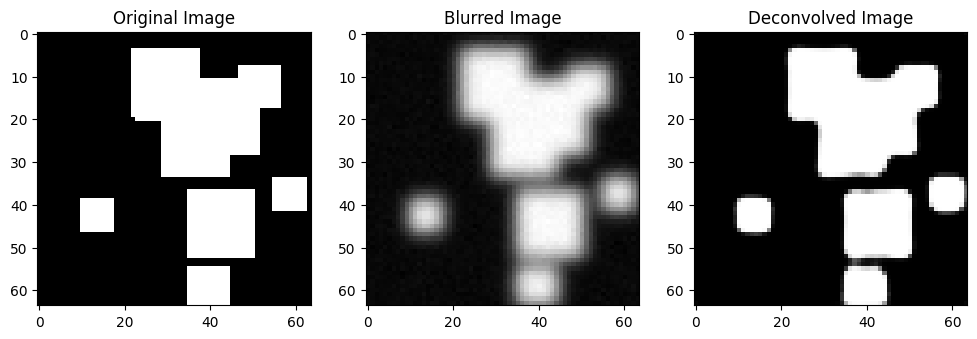}
\end{center}
\vspace{-12pt}
\caption{An example of synthetic IR image restoration: A synthetic example used for testing the trained NN. From left to right: Original, Blurred, Estimated.}
\end{figure} 

\begin{figure}[h!]
\begin{center}
\includegraphics[scale=.40]{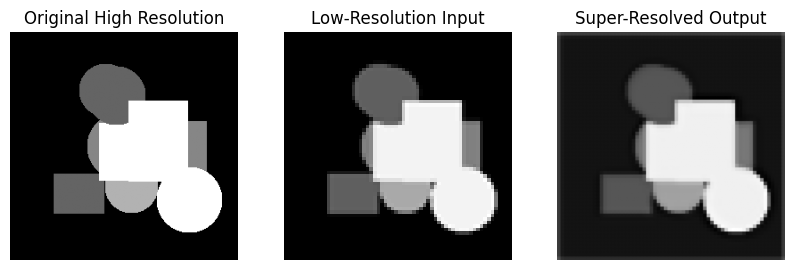}
\end{center}
\vspace{-12pt}
\caption{An example of synthetic IR image super-resolution: A synthetic example used for testing the trained NN. From left to right: Original high resolution (HR), Low resolution obtained by a downsampling of factor 2, Estimated.}
\end{figure} 

\begin{figure}[h!]
\begin{center}
\includegraphics[scale=.2]{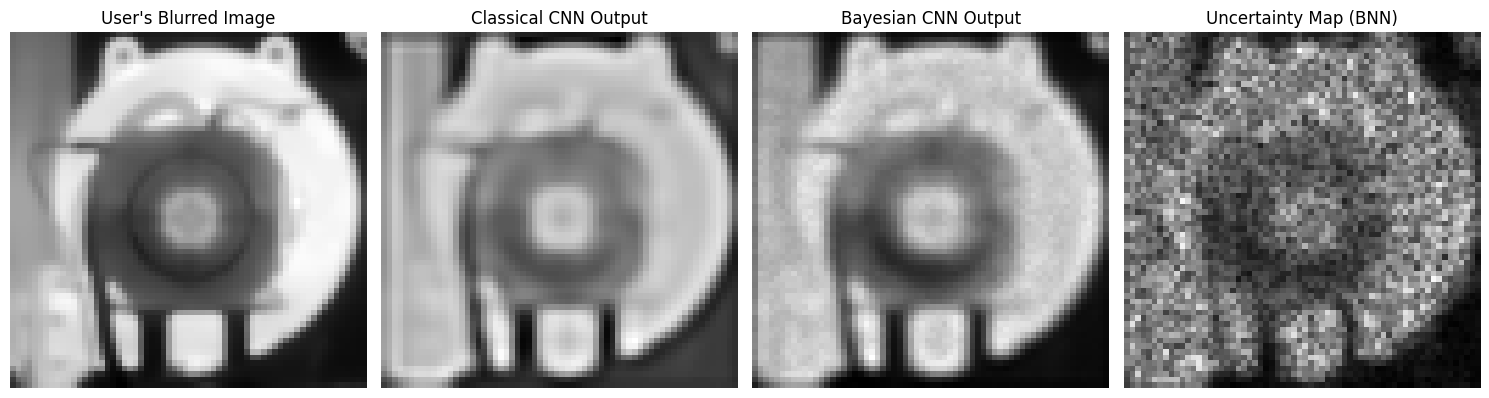}
\end{center}
\vspace{-12pt}
\caption{An example of IR image restoration: from left to right: Original, Blurred, Estimated by BPINN, mean and variances.}
\end{figure} 

\begin{figure}[h!]
\begin{center}
\includegraphics[scale=.35]{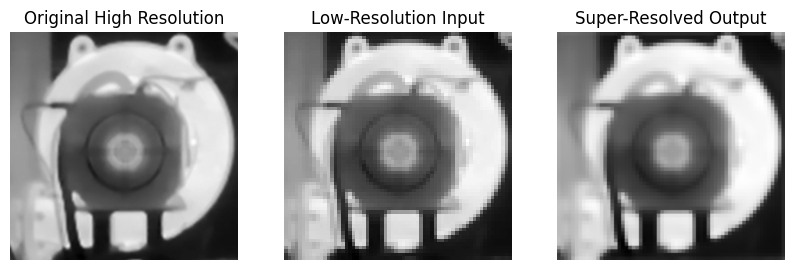}
\end{center}
\vspace{-12pt}
\caption{An example of IR image super-resolution: from left to right: original High Resolution (HR), Low resolution image at the input of the trained model, output of the model.}
\end{figure} 

\section{Conclusion}

In this paper, we introduced a Bayesian Physics-Informed Neural Network (BPINN) framework for solving linear inverse problems, called BPINN-IP, and with a specific focus on infrared image processing tasks like deconvolution and super-resolution. 

By integrating Bayesian inference into the PINN structure, we achieved, not only more robust parameter estimation, but also meaningful uncertainty quantification, which is crucial for some real-world applications with noisy and incomplete data. Of course, the Bayesian approach has been used  in different ways in NN based methods since many years. See for example \cite{Seeger2004,Blundell2015,Kendall2017,Sun2019,Rudner2023,Wenzel2020}. However, the proposed Bayesian approach in PINN is new, as we use a probabilist modeling in the generation or acquisition of the data set used forthe training using the forward model and accountion for the errors. 
In fact the Physics is used both in the training data set generation, and in training of the NN. The obtained posterior probabilities can be used for the training part and for the testing or inference part. 

Through simulation studies and real industrial examples, we demonstrated that BPINNs outperform traditional methods in both accuracy and robustness, while reducing reliance on large labeled datasets. The ability to encode prior knowledge and physical constraints directly into the learning process significantly enhances generalization capabilities, especially in ill-posed and high-dimensional settings.

Future work will explore extending the BPINN framework to handle nonlinear and dynamic inverse problems, optimizing network architectures for faster convergence, and validating the method on larger and more complex real-world datasets. With these improvements, we believe BPINNs can become a powerful, practical tool for solving a wide range of inverse problems across science and engineering.


\bibliographystyle{IEEEbib}
\bibliography{strings,refs}

\end{document}

%% file: macros_gpi.tex
\def\bdoc{\begin{document}}             \def\edoc{\end{document}}
\def\babs{\begin{abstract}}             \def\eabs{\end{abstract}}
\def\bcc{\begin{center}}                \def\ecc{\end{center}}
\def\ben{\begin{enumerate}}             \def\een{\end{enumerate}}
\def\bit{\begin{itemize}}               \def\eit{\end{itemize}}
\def\bpic{\begin{picture}}              \def\epic{\end{picture}}
\def\barr{\begin{array}}                \def\earr{\end{array}}
\def\btabu{\begin{tabular}}             \def\etabu{\end{tabular}}
\def\bcc{\begin{center}}                \def\ecc{\end{center}}
\def\bfig{\begin{figure}}               \def\efig{\end{figure}}
\def\bver{\begin{verb}}					\def\ever{\end{verb}}
\def\beq{\begin{equation}}				\def\eeq{\end{equation}}
\def\bmult{\begin{multline}}			\def\emult{\end{multline}}
\def\balign{\begin{align}}				\def\ealign{\end{align}}
\def\beqn{\begin{eqnarray}}             \def\eeqn{\end{eqnarray}}
\def\beqnx{\begin{eqnarray*}}           \def\eeqnx{\end{eqnarray*}}
\def\bbib{\begin{thebibliography}{999}}	\def\ebib{\end{thebibliography}}

\def\bm#1{\mbox{\boldmath $#1$}}

\def\zerob{{\bm 0}}
\def\oneb{{\bm 1}}

\def\ab{{\bm a}}
\def\bb{{\bm b}}
\def\cb{{\bm c}}
\def\db{{\bm d}}
\def\eb{{\bm e}}
\def\fb{{\bm f}}
\def\gb{{\bm g}}
\def\hb{{\bm h}}
\def\ib{{\bm i}}
\def\jb{{\bm j}}
\def\kb{{\bm k}}
\def\lb{{\bm l}}
\def\mb{{\bm m}}
\def\nb{{\bm n}}
\def\ob{{\bm o}}
\def\pb{{\bm p}}
\def\qb{{\bm q}}
\def\rb{{\bm r}}
\def\sb{{\bm s}}
\def\tb{{\bm t}}
\def\ub{{\bm u}}
\def\vb{{\bm v}}
\def\wb{{\bm w}}
\def\xb{{\bm x}}
\def\yb{{\bm y}}
\def\zb{{\bm z}}

\def\Ab{{\bm A}}
\def\Bb{{\bm B}}
\def\Cb{{\bm C}}
\def\Db{{\bm D}}
\def\Eb{{\bm E}}
\def\Fb{{\bm F}}
\def\Gb{{\bm G}}
\def\Hb{{\bm H}}
\def\Ib{{\bm I}}
\def\Jb{{\bm J}}
\def\Kb{{\bm K}}
\def\Lb{{\bm L}}
\def\Mb{{\bm M}}
\def\Nb{{\bm N}}
\def\Ob{{\bm O}}
\def\Pb{{\bm P}}
\def\Qb{{\bm Q}}
\def\Rb{{\bm R}}
\def\Sb{{\bm S}}
\def\Tb{{\bm T}}
\def\Ub{{\bm U}}
\def\Vb{{\bm V}}
\def\Wb{{\bm W}}
\def\Xb{{\bm X}}
\def\Yb{{\bm Y}}
\def\Zb{{\bm Z}}

\def\alphab{\bm{\alpha}}
\def\betab{\bm{\beta}}
\def\deltab{\bm{\delta}}
\def\etab{\bm{\eta}}
\def\epsilonb{\bm{\epsilon}}
\def\gammab{\bm{\gamma}}
\def\omegab{\bm{\omega}}
\def\etab{\bm{\eta}}
\def\thetab{\bm{\theta}}
\def\xib{\bm{\xi}}
\def\lambdab{\bm{\lambda}}
\def\taub{\bm{\tau}}
\def\phib{\bm{\phi}}
\def\mub{\bm{\mu}}
\def\psib{\bm{\psi}}
\def\chib{\bm{\chi}}
\def\sigmab{\bm{\sigma}}

\def\Deltab{\bm{\Delta}}
\def\Lambdab{\bm{\Lambda}}
\def\Phib{\bm{\Phi}}
\def\Psib{\bm{\Psi}}
\def\Sigmab{\bm{\Sigma}}

\def\Ac{{\cal A}}
\def\Bc{{\cal B}}
\def\Cc{{\cal C}}
\def\Dc{{\cal D}}
\def\Ec{{\cal E}}
\def\Fc{{\cal F}}
\def\Gc{{\cal G}}
\def\Hc{{\cal H}}
\def\Ic{{\cal I}}
\def\Jc{{\cal J}}
\def\Kc{{\cal K}}
\def\Lc{{\cal L}}
\def\Mc{{\cal M}}
\def\Nc{{\cal N}}
\def\Oc{{\cal O}}
\def\Pc{{\cal P}}
\def\Qc{{\cal Q}}
\def\Rc{{\cal R}}
\def\Sc{{\cal S}}
\def\Tc{{\cal T}}
\def\Uc{{\cal U}}
\def\Vc{{\cal V}}
\def\Wc{{\cal W}}
\def\Xc{{\cal X}}
\def\Yc{{\cal Y}}
\def\Zc{{\cal Z}}

\def\disp#1{\displaystyle{#1}}

\def\ra{\rightarrow}
\def\la{\leftarrow}
\def\da{\downarrow}
\def\ua{\uparrow}

\def\Ra{\Rightarrow}
\def\La{\Leftarrow}
\def\Da{\Downarrow}
\def\Ua{\Uparrow}

\def\lra{\longrightarrow}
\def\lla{\longleftarrow}
\def\Lra{\Longrightarrow}
\def\Lla{\longleftarrow}

\def\lrarr{\leftrightarrow}
\def\Lrarr{\Leftrightarrow}
\def\udarr{\updownarrow}
\def\Uparr{\Updownarrow}

\def\d#1{\,\mbox{d} #1}
\def\iii{\int_{-\infty}^{+\infty}}
\def\izi{\int_{0}^{\infty}}
\def\izpi{\int_{0}^{\pi}}
\def\izdpi{\int_{0}^{2\pi}}
\def\intd{\int\kern-.8em\int}
\def\intt{\int\kern-.8em\int\kern-.8em\int}
\def\intg{\int\kern-1.1em\int}

\def\wh{\widehat}
\def\xh{\widehat{x}}
\def\thetah{\widehat{\theta}}
\def\betah{\widehat{\beta}}

\def\xbh{\widehat{\xb}}
\def\ybh{\widehat{\yb}}
\def\thetabh{\widehat{\thetab}}
\def\etabh{\widehat{\etab}}
\def\betabh{\widehat{\betab}}

\def\xbhk{\widehat{\xb}^{k}}
\def\thetahk{\widehat{\theta}^{k}}
\def\betahk{\widehat{\beta}^{k}}

\def\xbhkp{\widehat{\xb}^{k+1}}
\def\thetahkp{\widehat{\theta}^{k+1}}
\def\betahkp{\widehat{\beta}^{k+1}}

\def\thetabhk{\widehat{\thetab}^{k}}
\def\betabhk{\widehat{\betab}^{k}}

\def\thetabhkp{\widehat{\thetab}^{k+1}}
\def\betabhkp{\widehat{\betab}^{k+1}}

\def\thetamin{\theta_{\mbox{\tiny min}}}
\def\thetamax{\theta_{\mbox{\tiny max}}}
\def\betamin{\beta_{\mbox{\tiny min}}}
\def\betamax{\beta_{\mbox{\tiny max}}}

\def\expf#1{\exp\left[ {#1} \right]}

\def\argmax#1#2{\arg\max_{#1}\left\{ #2 \right\}}
\def\argmin#1#2{\arg\min_{#1}\left\{ #2 \right\}}

\def\Prob#1{\mbox{Pr}\left\{#1\right\}}
\def\var#1{\mbox{Var}\left\{#1\right\}}
\def\cov#1{\mbox{Cov}\left\{#1\right\}}
\def\corr#1{\mbox{Corr}\left\{#1\right\}}
\def\rang#1{\mbox{rang}\left\{#1\right\}}
\def\det#1{\mbox{d\'et}\left\{#1\right\}}

\def\gbu{\underline{\gb}}
\def\fbu{\underline{\fb}}
\def\zbu{\underline{\zb}}
\def\epsilonbu{\underline{\epsilonb}}
\def\thetabu{\underline{\thetab}}

\def\ve{{v_{\epsilon}}}
\def\sigmae{{\sigma_{\epsilon}}}
\def\Sigmae{{\Sigmab_{\epsilonb}}}
\def\Sigmaf{{\Sigmab_{\fb}}}
\def\Sigmag{{\Sigmab_{\gb}}}

\def\fbuh{\widehat{\fbu}}
\def\zbuh{\widehat{\fbu}}
\def\thetabuh{\widehat{\thetabu}}

\def\fbh{\widehat{\fb}}
\def\Pbh{\widehat{\Pb}}
\def\Sigmabh{\widehat{\Sigmab}}
\def\Abh{\widehat{\Ab}}
\def\Bbh{\widehat{\Bb}}
\def\thetabh{\widehat{\thetab}}
\def\Rbe{\Rb_{\epsilon}}
\def\Rbeh{\widehat{\Rbe}}

\def\Rjk{{R_j}_k}
\def\fbik{{\fb_i}_k}
\def\fbjk{{\fb_j}_k}

\def\mik{{m_i}_k}
\def\sigmaik{{\sigma_i^2}_k}
\def\Sigmaik{{\Sigmab_i}_k}
\def\alphaik{{\alpha_i}_k}
\def\betaik{{\beta_i}_k}

\def\muik{{\mu_i}_k}
\def\vik{{v_i}_k}

\def\mjk{{m_j}_k}
\def\sigmajk{{\sigma_j^2}_k}
\def\Sigmajk{{\Sigmab_j}_k}
\def\alphajk{{\alpha_j}_k}
\def\betajk{{\beta_j}_k}

\def\mujk{{\mu_j}_k}
\def\vjk{{v_j}_k}
\def\njk{{n_j}_k}

\def\alphae{\alpha_{\epsilon}}
\def\betae{\beta_{\epsilon}}
\def\alphaez{\alpha_{\epsilon_0}}
\def\betaez{\beta_{\epsilon_0}}
\def\alphaei{\alpha_{\epsilon_i}}
\def\betaei{\beta_{\epsilon_i}}
\def\alphaeiz{\alpha_{\epsilon_{i_0}}}
\def\betaeiz{\beta_{\epsilon_{i-0}}}

\def\mbz{{{\mb}_z}}
\def\Sigmabz{{{\Sigmab}_z}}
\def\diag#1{\mbox{diag}\left[#1\right]}

\def\sigmah{\widehat{\sigma}}
\def\fh{\widehat{f}}
\def\sigmaz{\sigma_z}

\def\sumi{\sum_{i=1}^{M} }
\def\sumj{\sum_{j=1}^{N} }
\def\lambdah{\wh{\lambda}}
\def\lambdabh{\wh{\lambdab}}

\def\det#1{\hbox{det}\left(#1\right)}
\def\defined{\,\shortstack{$\triangle$\\ =}\,}

\def\zmat{\left[\matrix{0}\right]}
\def\det#1{\hbox{det}(#1)}
\def\th{\widehat{t}}
\def\xh{\widehat{x}}
\def\lambdah{\widehat{\lambda}}
\def\muh{\widehat{\mu}}
\def\sigmah{\widehat{\sigma}}
\def\rhoh{\widehat{\rho}}
\def\betah{\widehat{\beta}}
\def\alphah{\widehat{\alpha}}
\def\tauh{\widehat{\tau}}

\def\tbh{\widehat{\tb}}
\def\mubh{\widehat{\mub}}
\def\sigmabh{\widehat{\sigmab}}
\def\rhobh{\widehat{\rhob}}
\def\oneb{\mbox{\bf 1}}

\def\bm#1{\mbox{\boldmath $#1$}}
\def\zerob{{\bf 0}}
\def\oneb{{\bf 1}}
\def\intg{\int\kern-1.1em\int}
\def\expf#1{\exp\left[ {#1} \right]}

\def\gbu{\underline{\gb}}
\def\fbu{\underline{\fb}}
\def\zbu{\underline{\zb}}
\def\epsilonbu{\underline{\epsilonb}}
\def\uthetab{\underline{\thetab}}

\def\sigmae{{\sigma_{\epsilon}}}
\def\Sigmae{{\Sigma_{\epsilonb}}}
\def\Sigmabe{{\Sigmab_{\epsilonb}}}
\def\Sigmaf{{\Sigmab_{\fb}}}
\def\Sigmag{{\Sigmab_{\gb}}}

\def\fbuh{\widehat{\fbu}}
\def\zbuh{\widehat{\fbu}}
\def\uthetabh{\widehat{\uthetab}}

\def\fbh{\widehat{\fb}}
\def\Sigmabh{\widehat{\Sigmab}}
\def\Abh{\widehat{\Ab}}
\def\thetabh{\widehat{\thetab}}

\def\Rjk{{R_j}_k}
\def\fbik{{\fb_i}_k}
\def\fbjk{{\fb_j}_k}

\def\mik{{m_i}_k}
\def\sigmaik{{\sigma_i^2}_k}
\def\Sigmaik{{\Sigmab_i}_k}
\def\alphaik{{\alpha_i}_k}
\def\betaik{{\beta_i}_k}

\def\muik{{\mu_i}_k}
\def\vik{{v_i}_k}

\def\mjk{{m_j}_k}
\def\sigmajk{{\sigma_j^2}_k}
\def\Sigmajk{{\Sigmab_j}_k}
\def\alphajk{{\alpha_j}_k}
\def\betajk{{\beta_j}_k}

\def\mujk{{\mu_j}_k}
\def\vjk{{v_j}_k}
\def\njk{{n_j}_k}

\def\mbz{{{\mb}_z}}
\def\Sigmabz{{{\Sigmab}_z}}

\def\vect{\mbox{vect}}
\def\lra{\longrightarrow}

\def\gir{g_i(\rb)}
\def\fjr{f_j(\rb)}
\def\zjr{z_j(\rb)}
\def\epsilonir{\epsilon_i(\rb)}
\def\gbr{\gb(\rb)}
\def\fbr{\fb(\rb)}
\def\zbr{\zb(\rb)}
\def\epsilonbr{\epsilonb(\rb)}
\def\fbhr{\fbh(\rb)}
\def\Sigmabhr{\Sigmabh(\rb)}

\def\mbzr{\mb_{z(\rb)}}
\def\Sigmabzr{\Sigmab_{z(\rb)}}
\def\Sigmagbzr{{\Sigmab_g}_{z(\rb)}}

\def\mjzjr#1{{m_{#1}}_{z_{#1}(\rb)}}
\def\sigmajzjr#1{{\sigma_{#1}}_{z_{#1}(\rb)}}

\def\Beta{B}
\def\Betab{\bm{\Beta}}
\def\Betabe{\bm{\Beta}_{\epsilonb}}

\def\fh{\widehat{f}}
\def\sigmah{\widehat{\sigma}}
\def\sigmaei{\sigma_{\epsilon_i}^2}

\def\thetah{\widehat{\theta}}
\def\sigmah{\widehat{\sigma}}

\def\Ftxm{\widetilde{F}_{X}(x_-)}
\def\Ftxp{\widetilde{F}_{X}(x_+)}

\def\NU{{\cal V}}

\def\tFx{\widetilde{F}_{X}}

\def\Fx{F_{X}(x)}
\def\fx{f_{X}(x)}

\def\Fxv{F_{X}(x)}
\def\fxv{f_{X}(x)}

\def\Fxi{{F}_{X_i}(x_i)}
\def\fxi{{f}_{X_i}(x_i)}

\def\Gnu{G_{\NU}(\nu)}
\def\gnu{g_{\NU}(\nu)}

\def\Fnu{F_{\NU}(\nu)}
\def\fnu{f_{\NU}(\nu)}

\def\Finnu#1{F^{-1}_{\NU}(#1)}

\def\Gnua#1{G_{\NU}(#1)}
\def\Gnub#1{G_{\NU}^{-1}(#1)}
\def\Gnuh{G_{\NU}^{-1}(1/2)}

\def\Ftx{\widetilde{F}_{X}(x)}
\def\ftx{\widetilde{f}_{X}(x)}

\def\Ftxi{\widetilde{F}_{X_i}(x_i)}
\def\ftxi{\widetilde{f}_{X_i}(x_i)}

\def\Ftxv{\widetilde{F}_{X}(x)}
\def\FtX#1{\widetilde{F}_{X}(#1)}
\def\ftxv{\widetilde{f}_{X}(x)}

\def\fxnu{f_{X|\NU}(x|\nu)}
\def\fxNU{f_{X|\NU}(x|\NU)}
\def\FxNU{F_{X|\NU}(x|\NU)}

\def\fxNUtheta{f_{X|\NU,\theta}(x|\NU,\theta)}
\def\FxNUtheta{F_{X|\NU,\theta}(x|\NU,\theta)}

\def\Fxnu{F_{X|\NU}(x|\nu)}
\def\Fxnui#1{F_{X|\NU}(x|\nu_#1)}

\def\FxN#1{F_{X|\NU}(#1)}
\def\FxNU{F_{X|\NU}(x|\NU)}
\def\Fxnun#1{F_{X|\NU}(x|#1)}
\def\Fxinun#1{F_{X_i|\NU}(x_i|#1)}
\def\fxnu{f_{X|\NU}(x|\nu)}
\def\fxNU{f_{X|\NU}(x|\NU)}

\def\Fxvnu{F_{X|\NU}(x|\nu)}
\def\Fxvnun#1{F_{X|\NU}(x|#1)}
\def\FxvNU{F_{X|\NU}(x|\NU)}

\def\FXvnu{F_{X|\NU}}

\def\fxvnu{f_{X|\NU}(x|\nu)}
\def\fxvNU{f_{X|\NU}(x|\NU)}

\def\FXvNU#1{{F}_{X|\NU}(#1|\NU)}
\def\FXvn#1{{F}_{X|\NU}(x|#1)}

\def\Fxtheta{F_{X|\theta}(x|\theta)}
\def\fxtheta{f_{X|\theta}(x|\theta)}

\def\Fxvtheta{F_{X|\theta}(x|\theta)}
\def\fxvtheta{f_{X|\theta}(x|\theta)}

\def\Ftxtheta{\widetilde{F}_{X|\theta}(x|\theta)}
\def\ftxtheta{\widetilde{f}_{X|\theta}(x|\theta)}

\def\Ftxvtheta{\widetilde{F}_{X|\theta}(x|\theta)}
\def\ftxvtheta{\widetilde{f}_{X|\theta}(x|\theta)}

\def\Fxnutheta{F_{X|\NU,\theta}(x|\nu,\theta)}
\def\fxnutheta{f_{X|\NU,\theta}(x|\nu,\theta)}

\def\ftheta{f_{\Theta}(\theta)}
\def\fthetaxnuz{f_{\Theta|X,\nu_0}(\theta|x,\nu_0)}
\def\fthetax{f_{\Theta|X}(\theta|x)}

\def\Fxvnutheta{F_{X|\NU,\theta}(x|\nu,\theta)}
\def\FxvNUtheta{F_{X|\NU,\theta}(x|\NU,\theta)}
\def\fxvnutheta{f_{X|\NU,\theta}(x|\nu,\theta)}
\def\Ftxvnutheta{\widetilde{F}_{X|\theta}(x|\theta)}
\def\ftxvnutheta{\widetilde{f}_{X|\theta}(x|\theta)}

\def\FxvNUm{F_{X|\NU}(x_-|\NU)}
\def\FxvNUp{F_{X|\NU}(x_+|\NU)}

\def\Fxinu{F_{X_i|\NU}(x_i|\nu)}
\def\fxinu{f_{X_i|\NU}(x_i|\nu)}

\def\Ftxi{\widetilde{F}_{X_i}(x_i)}
\def\ftxi{\widetilde{f}_{X_i}(x_i)}

\def\fxitheta{f_{X_i|\theta}(x_i|\theta)}
\def\ftxitheta{\widetilde{f}_{X_i|\theta}(x_i|\theta)}

\def\thetah{\widehat{\theta}}
\def\thetat{\widetilde{\theta}}

\def\lnuxv{l_{\nu|X}(\nu|x)}

\def\Lnuxva#1{L_{\nu|X}(#1 | x)}
\def\lnuxva#1{l_{\nu|X}(#1 | x)}

\def\Prob#1{\mbox{P}\left(#1\right)}

\def\lthetaxvnuz{l(\theta)}

\def\lthetaxvnuz{l_{\theta|x,\nu_0}(\theta | x,\nu_0)}
\def\fxnuztheta{f_{X|\nu_0,\theta}(x | \nu_0,\theta)}
\def\fxinuztheta{f_{X|\nu_0,\theta}(x_i | \nu_0,\theta)}
\def\fxvtheta{f_{X|\theta}(x | \theta)}
\def\lthetaxv{l_{\theta|X}(\theta | x)}
\def\ltthetaxv{\widetilde{l}_{\theta|X}(\theta | x)}

\def\ltheta{l(\theta )}
\def\lttheta{\widetilde{l}(\theta)}
\def\Lnu{L(\nu )}
\def\Lnui#1{L(\nu_#1)}
\def\Linu{L_i(\nu )}
\def\Linu{L_i(\nu )}
\def\LNU{L(\NU )}
\def\Lin#1{L^{-1}(#1)}
\def\L#1{L(#1)}
\def\Li#1{L_i(#1)}
\def\FNU#1{F_\NU(#1)}
\def\FinNU#1{F^{-1}_\NU(#1)}

\def\ftthetax{\tilde{f}_{\Theta|X}(\theta|x)}
\def\ER{R}

\def\Ndd#1#2#3{{\cal N}\left( #1 ;~ #2, #3 \right)}
\def\Cdd#1#2#3{{\cal C}\left( #1 ;~ #2, #3 \right)}
\def\Sdd#1#2#3#4{{\cal S}\left( #1 ;~ #2, #3, #4 \right)}
\def\Edd#1#2{{\cal E}\left( #1 ;~ #2 \right)}
\def\DEdd#1#2{{\cal DE}\left( #1 ;~ #2 \right)}
\def\Gdd#1#2#3{{\cal G}\left( #1 ;~ #2, #3 \right)}
\def\IGdd#1#2#3{{\cal IG}\left( #1 ;~ #2, #3 \right)}

\def\Xwr{X(\omega,\rb)}
\def\xwr{x(\omega,\rb)}

\def\muzw{\mu_{z}(\omega)}
\def\muzwr{\mu_{z(\rb)}(\omega)}
\def\szw{\sigma_{z}^2(\omega)}
\def\szwr{\sigma_{z(\rb)}^2(\omega)}
\def\pzw{p_{z}(\omega)}
\def\pzwr{p_{z(\rb)}(\omega)}

\def\hszw{\widehat{\sigma_{z}^2}(\omega)}
\def\hmuzw{\widehat{\mu}_{z}(\omega)}
\def\hpzw{\widehat{p}_{z}(\omega)}

\def\muzwrl{\mu_{z(\rb)}(\omega-l)}

\def\pxwrcz{p\left(\xwr ~|~ z\right)}
\def\pxwr{p\left(\xwr\right)}

\def\Xbw{\Xb(\omega)}
\def\xbw{\xb(\omega)}
\def\uXb{\underline{\Xb}}
\def\uxb{\underline{\xb}}
\def\pxbw{p(\xbw)}
\def\puxb{p(\uxb)}
\def\pzb{P(\zb)}

\def\Szw{S_z(\omega)}
\def\mubw{\mub(\omega)}

\def\Srw{S_{\rb}(\omega)}
\def\Swr{S_{\omega}(\rb)}
\def\Sbw{\Sb(\omega)}
\def\sbw{\sb(\omega)}
\def\Sbr{\Sb(\rb)}
\def\mubw{\mub(\omega)}
\def\cov#1{\mbox{cov}[#1]}

\def\pdf{\emph{pdf}}
\def\pmf{\emph{pmf}}
\def\dpdx#1#2{\frac{\partial #1}{\partial #2}}

\def\REM#1{}

\def\XXwr{X(\omega,\rb)}
\def\xxwr{x(\omega,\rb)}

\def\Xwr{X_{\omega}(\rb)}
\def\xwr{x_{\omega}(\rb)}
\def\Xrw{X_{\rb}(\omega)}
\def\xrw{x_{\rb}(\omega)}

\def\Ywr{Y_{\omega}(\rb)}
\def\ywr{y_{\omega}(\rb)}
\def\Yrw{Y_{\rb}(\omega)}
\def\yrw{y_{\rb}(\omega)}

\def\Ewr{E_{\omega}(\rb)}
\def\ewr{\epsilon_{\omega}(\rb)}
\def\Erw{E_{\rb}(\omega)}
\def\erw{\epsilon_{\rb}(\omega)}

\def\Xbr{\Xb(\rb)}
\def\xbr{\xb(\rb)}
\def\Xbw{\Xb(\omega)}
\def\xbw{\xb(\omega)}

\def\uXb{\underline{\Xb}}
\def\uxb{\underline{\xb}}
\def\uXbz{\underline{\Xb}_z}
\def\uxbz{\underline{\xb}_z}

\def\uthetab{\underline{\thetab}}
\def\uthetabz{\underline{\thetab}_z}

\def\Ywr{Y_{\omega}(\rb)}
\def\ywr{y_{\omega}(\rb)}
\def\Yrw{Y_{\rb}(\omega)}
\def\yrw{y_{\rb}(\omega)}
\def\Ybr{\Yb(\rb)}
\def\ybr{\yb(\rb)}
\def\Ybw{\Yb(\omega)}
\def\ybw{\yb(\omega)}
\def\uYb{\underline{\Yb}}
\def\uyb{\underline{\yb}}
\def\uYbz{\underline{\Yb}_z}
\def\uybz{\underline{\yb}_z}

\def\Ewr{E_{\omega}(\rb)}
\def\ewr{\epsilon_{\omega}(\rb)}
\def\Erw{E_{\rb}(\omega)}
\def\erw{\epsilon_{\rb}(\omega)}
\def\Ebr{Eb(\rb)}
\def\ebr{\epsilonb(\rb)}
\def\Ebw{Eb(\omega)}
\def\ebw{\epsilonb(\omega)}
\def\uEb{\underline{\Eb}}

\def\Erwp{E_{\rb}(\omega')}
\def\Xrwp{X_{\rb}(\omega')}

\def\xwmr{x_{\omega-1}(\rb)}
\def\muzwm{\mu_{z}(\omega-1)}

\def\rhoz{\rho_z}
\def\sez{\sigma_{\eta_z}^2}

\def\xwrp{x_{\omega}(\rb')}
\def\xrwp{x_{\rb}(\omega')}
\def\xwpr{x_{\omega'}(\rb)}
\def\xwprp{x_{\omega'}(\rb')}
\def\Xwrp{x_{\omega}(\rb')}
\def\Xwrp{X_{\omega}(\rb')}
\def\Xwpr{X_{\omega'}(\rb)}
\def\Xwprp{X_{\omega'}(\rb')}

\def\sigmaed{\sigmae^2}
\def\sigmaew{\sigmae(\omega)}
\def\sigmaedw{\sigmaed(\omega)}

\def\Sigmabe{\Sigmab_{\epsilon}}
\def\sigmabh{\widehat{\Sigmab}}
\def\xbh{\widehat{\xb}}
\def\zbh{\widehat{\zb}}
\def\qbh{\widehat{\qb}}
\def\sbh{\widehat{\sb}}

\def\espx#1#2{\mbox{E}_{#1}\left\{#2\right\}}
\def\esp#1{\mbox{E}\left\{#1\right\}}
\def\d#1{\mbox{~d}#1}

\def\Tr#1{\mbox{Tr}\left[#1\right]}

\def\pmatrix#1{\left[\barr{ccccccccccccccc} #1 \earr\right]}
\def\Bf{\bm{B}}
\def\gbh{\widehat{\gb}}
\def\dbh{\widehat{\db}}
\def\Tr#1{\mbox{Tr}\left[#1\right]}
\def\Cov#1{\mbox{Cov}\left[#1\right]}
\def\iid{i.i.d.~}

\def\cro#1{\left[#1\right]}
\def\acc#1{\left\{#1\right\}}
\def\aprio{\emph{a priori~}}
\def\apost{\emph{a posteriori~}}
\def\hbh{\widehat{\hb}}
\def\qh{\widehat{q}}
\def\fbt{\widetilde{\fb}}
\def\thetabt{\widetilde{\thetab}}

\def\mubt{\widetilde{\mub}}
\def\Sigmabt{\widetilde{\Sigmab}}
\def\qt{\widetilde{q}}
\def\vt{\widetilde{v}}
\def\zt{\widetilde{z}}
\def\Dbt{\widetilde{\Db}}
\def\alphat{\widetilde{\alpha}}
\def\betat{\widetilde{\beta}}
\def\abt{\widetilde{\ab}}
\def\ajt{\widetilde{a}_j}
\def\taut{\widetilde{\tau}}
\def\mut{\widetilde{\mu}}
\def\Zbt{\widetilde{\Zb}}
\def\Abt{\widetilde{\Ab}}
\def\zbt{\widetilde{\zb}}

\def\bloca#1#2#3{
\begin{picture}(50,50)
  \put(32,42){\makebox(0,0){#1}}
  \put(25,50){\vector(0,-1){20}}
  \put(0,10){\framebox(50,20){#2}}
  \put(0,0){\makebox(50,0){#3}}  
\end{picture}
}

\def\ugb{\underline{\gb}}
\def\fbh{\widehat{\fb}}
\def\zbh{\widehat{\zb}}
\def\qbh{\widehat{\qb}}
\def\thetabh{\widehat{\thetab}}
\def\Pbh{\widehat{\Pb}}

\def\sigmae{\sigma_{\epsilon}}
\def\REM#1{}
\def\fh{\widehat{f}}
\def\zh{\widehat{z}}
\def\disp{\displaystyle}
\def\oneb{{\boldmath{1}}}
\def\zerob{{\boldmath{0}}}
\def\d#1{\; \mbox{d} #1}
\def\expf#1{\exp\left\{#1\right\}}
\def\esp#1{\mbox{E}\left\{#1\right\}}
\def\lra{\longrightarrow}

\def\bslide#1{\begin{frame}{\Large\bf #1}}
\def\eslide{\end{frame}}

\def\inversions#1#2#3{
\begin{picture}(100,40)
  \put(0,15){\makebox(15,0){#1}}
  \put(20,15){\vector(1,0){10}}
  \put(30,7){\framebox(40,15){#2}}
  \put(70,15){\vector(1,0){10}}
  \put(85,15){\makebox(15,0){#3}}
\end{picture}
}

\def\blue#1{{\color{blue}#1}}
\def\red#1{{\color{red}#1}}
\def\green#1{{\color{green}#1}}

\def\rfb{\red{\fb}}
\def\bgb{\blue{\gb}}
\def\rfbh{\red{\widehat{\fb}}}
\def\rfbt{\red{\widetilde{\fb}}}
\def\rzbt{\red{\widetilde{\zb}}}
\def\rqb{\red{\qb}}
\def\rqbh{\red{\widehat{\qb}}}
\def\rzb{\red{\zb}}
\def\rzbh{\red{\widehat{\zb}}}
\def\rthetab{\red{\thetab}}
\def\retab{\red{\etab}}
\def\rthetabh{\red{\widehat{\thetab}}}
\def\rthetabt{\red{\widetilde{\thetab}}}
\def\retabt{\red{\widetilde{\etab}}}
\def\ralphabt{\red{\widetilde{\alphab}}}

\def\ugb{\underline{\gb}}
\def\fbh{\widehat{\fb}}
\def\zbh{\widehat{\zb}}
\def\qbh{\widehat{\qb}}
\def\thetabh{\widehat{\thetab}}
\def\Pbh{\widehat{\Pb}}

\def\UP#1{\uppercase{#1}}
\def\sca#1{\uppercase{#1}}

\def\sigmae{\sigma_{\epsilon}}
\def\REM#1{}
\def\fh{\widehat{f}}
\def\zh{\widehat{z}}
\def\disp{\displaystyle}
\def\oneb{{\boldmath{1}}}
\def\zerob{{\boldmath{0}}}
\def\d#1{\; \mbox{d} #1}
\def\expf#1{\exp\left\{#1\right\}}
\def\esp#1{\mbox{E}\left\{#1\right\}}
\def\lra{\longrightarrow}

\def\rfb{\red{\fb}}
\def\rFb{\red{\Fb}}
\def\bgb{\blue{\gb}}
\def\rfbh{\red{\widehat{\fb}}}
\def\rqb{\red{\qb}}
\def\rqbh{\red{\widehat{\qb}}}
\def\rzb{\red{\zb}}
\def\rzbh{\red{\widehat{\zb}}}
\def\rthetab{\red{\thetab}}
\def\ralphab{\red{\alphab}}
\def\rbetab{\red{\betab}}
\def\rthetabh{\red{\widehat{\thetab}}}
\def\ralphabh{\red{\widehat{\alphaab}}}
\def\rSigmab{\red{\Sigmab}}
\def\rSigmabe{\red{\Sigmabe}}

\def\rz{\red{z}}
\def\rzh{\red{\widehat{z}}}

\def\rab{\red{\ab}}
\def\rabh{\red{\widehat{\ab}}}
\def\rAb{\red{\Ab}}
\def\rAbh{\red{\widehat{\Ab}}}
\def\rBb{\red{\Bb}}
\def\rBbh{\red{\widehat{\Bb}}}
\def\rPbh{\red{\widehat{\Pb}}}
\def\rub{\red{\ub}}
\def\rubh{\red{\widehat{\ub}}}

\def\rVb{\red{\Vb}}
\def\rVbh{\widehat{\rVb}}
\def\rVbt{\widetilde{\rVb}}

\def\rvb{\red{\vb}}
\def\rvbh{\widehat{\rvb}}
\def\rvbt{\widetilde{\rvb}}

\def\bfb{\blue{\fb}}
\def\bFb{\blue{\Fb}}

\def\rfi{\red{f}_i}
\def\rfj{\red{f}_j}
\def\ruj{\red{u}_j}
\def\raj{\red{a}_j}
\def\rzj{\red{z}_j}
\def\rvj{\red{v}_j}
\def\rtau{\red{\tau}}
\def\rtauj{\red{\tau_j}}
\def\rtaub{\red{\taub}}
\def\rtauh{\widehat{\rtau}}
\def\rtaujh{\widehat{\rtauj}}
\def\rtaubh{\widehat{\rtaub}}
\def\rve{\red{v_{\epsilon}}}
\def\rv{\red{v}}

\def\rfjh{\red{\widehat{f}}_j}
\def\rujh{\red{\widehat{u}}_j}
\def\rajh{\red{\widehat{a}}_j}
\def\rzjh{\red{\widehat{z}}_j}
\def\rvjh{\red{\widehat{v}}_j}

\def\ralpha{\red{\alpha}}
\def\rbeta{\red{\beta}}
\def\rtheta{\red{\theta}}

\def\rlambda{\red{\lambda}}
\def\ralphah{\widehat{\ralpha}}
\def\rbetah{\widehat{\rbeta}}
\def\rthetah{\widehat{\rtheta}}
\def\rlambdah{\widehat{\rlambda}}

\def\ubh{\widehat{\ub}}

\def\rve{\red{v_{\epsilon}}}
\def\alphae{\alpha_{\epsilon_0}}
\def\betae{\beta_{\epsilon_0}}
\def\alphajh{\widehat{\alpha}_j}
\def\betajh{\widehat{\beta}_j}
\def\alphaeh{\widehat{\alpha}_{\epsilon_0}}
\def\betaeh{\widehat{\beta}_{\epsilon_0}}

\def\rqb{\red{\qb}}
\def\rqbh{\red{\widehat{\qb}}}
\def\rzb{\red{\zb}}
\def\rZb{\red{\Zb}}
\def\rab{\red{\ab}}
\def\rAb{\red{\Ab}}
\def\rvb{\red{\vb}}
\def\rmb{\red{\mb}}
\def\rzbh{\red{\widehat{\zb}}}
\def\rvbh{\red{\widehat{\vb}}}
\def\rmbh{\red{\widehat{\mb}}}
\def\rthetab{\red{\thetab}}
\def\rthetabh{\red{\widehat{\thetab}}}
\def\rfbh{\widehat{\rfb}}
\def\rhb{\red{\hb}}
\def\rhbh{\widehat{\rhb}}
\def\rthetai{\red{\theta}_i}
\def\rthetab{\red{\thetab}}
\def\rfjm{\red{f}_{j-1}}
\def\rfim{\red{f}_{i-1}}

\def\muh{\widehat{\mu}}
\def\vh{\widehat{v}}

\def\rSigmabh{\red{\Sigmabh}}
\def\rVbh{\red{\Vbh}}
\def\Fbh{\red{\Fb}}
\def\rFbh{\red{\Fbh}}

\def\Sigmabeh{\widehat{\Sigmabe}}
\def\rSigmabeh{\widehat{\red{\Sigmabe}}}

\def\rf{\red{f}}
\def\bg{\blue{g}}
\def\rF{\red{F}}
\def\bG{\blue{G}}

\def\sumk{\sum_{k=1}^K}
\def\suml{\sum_{l=1}^L}
\def\sumn{\sum_{n=1}^N}
\def\sumj{\sum_{j=1}^J}
\def\sumi{\sum_{i=1}^I}

\def\nub{\bm{\nu}}
\def\rak{\red{a_k}}
\def\rbk{\red{b_k}}
\def\rck{\red{c_k}}
\def\rtk{\red{t_k}}
\def\rvk{\red{v_k}}
\def\rnuk{\red{\nu_k}}
\def\rnuz{\red{\nu_0}}
\def\rphik{\red{\phi_k}}
\def\rnub{\red{\nub}}
\def\rtb{\red{\tb}}
\def\rphib{\red{\phib}}
\def\rfn{\red{f_n}}
\def\rtn{\red{t_n}}
\def\rvn{\red{v_n}}

\def\rve{\red{v_\epsilon}}
\def\rvf{\red{v_f}}
\def\rveh{\red{\widehat{v}_\epsilon}}
\def\rvfh{\red{\widehat{v}_f}}

\def\rnu{\red{\nu}}
\def\rphi{\red{\phi}}
\def\rbl{\red{b_l}}
\def\rbb{\red{\bb}}

\def\modela{\tt\setlength{\unitlength}{0.92pt}
\begin{picture}(250,106)
\thinlines    \put(-5,105){$\Nc(t|\red{t_1,v_1})$}
              \put(35,100){\vector(1,0){25}}
              \put(60,95){\framebox(37,16){\red{$a_1$}}}
			  \multiput(77,72)(0,7){3}{\circle*{4}}
              \put(-5,65){$\Nc(t|\red{t_k,v_k})$}
              \put(35,59){\vector(1,0){25}}
              \put(60,51){\framebox(37,16){\red{$a_k$}}}
              \multiput(77,32)(0,7){3}{\circle*{4}}              
              \put(-5,23){$\Nc(t|\red{t_K,v_K})$}
              \put(60,10){\framebox(37,16){\red{$a_K$}}}
              \put(35,18){\vector(1,0){25}}

              \put(98,103){\vector(1,-1){38}}
              \put(98,59){\vector(1,0){39}}
              \put(98,17){\vector(1,1){38}}

              \put(150,60){\circle{24}}
              \put(146,57){$\Large+$}
              \put(165,75){$g_{\rthetab}(t)$}
              \put(164,60){\vector(1,0){33}}
              
              \put(212,85){$\epsilon(t)$}
              \put(210,95){\vector(0,-1){20}}
              \put(211,60){\circle{24}}
              \put(207,58){$\Large+$}
              
              \put(225,66){$g(t)$}
              \put(225,60){\vector(1,0){20}}
\end{picture}
}

\def\modelb{\tt\setlength{\unitlength}{0.92pt}
\begin{picture}(250,106)
\thinlines    \put(-5,105){$\delta(t-\red{t_1})$}
              \put(0,100){\vector(1,0){37}}
              \put(38,95){\framebox(60,16){$\red{a_1} \Nc(t|0,\red{v_1})$}}
			  \multiput(77,72)(0,7){3}{\circle*{4}}
              \put(-5,65){$\delta(t-\red{t_k})$}
              \put(0,59){\vector(1,0){37}}
              \put(38,51){\framebox(60,16){$\red{a_k} \Nc(t|0,\red{v_k})$}}
              \multiput(77,32)(0,7){3}{\circle*{4}}              
              \put(-5,23){$\delta(t-\red{t_K})$}
              \put(38,10){\framebox(60,16){$\red{a_K} \Nc(t|0,\red{v_K})$}}
              \put(0,18){\vector(1,0){37}}

              \put(98,103){\vector(1,-1){38}}
              \put(98,59){\vector(1,0){39}}
              \put(98,17){\vector(1,1){38}}

              \put(150,60){\circle{24}}
              \put(146,57){$\Large+$}
              \put(165,75){$g_{\rthetab}(t)$}
              \put(164,60){\vector(1,0){33}}
              
              \put(212,85){$\epsilon(t)$}
              \put(210,95){\vector(0,-1){20}}
              \put(211,60){\circle{24}}
              \put(207,58){$\Large+$}
              
              \put(225,66){$g(t)$}
              \put(225,60){\vector(1,0){20}}
\end{picture}
}

\def\modelc{\tt\setlength{\unitlength}{0.92pt}
\begin{picture}(250,106)
\thinlines    \put(-5,105){$\delta(t-\red{t_1})$}
              \put(0,100){\vector(1,0){37}}
              \put(38,95){\framebox(40,16){$\red{a_1}$}}
			  \multiput(57,72)(0,7){3}{\circle*{4}}
              \put(-5,65){$\delta(t-\red{t_k})$}
              \put(0,59){\vector(1,0){37}}
              \put(38,51){\framebox(40,16){$\red{a_k}$}}
              \multiput(57,32)(0,7){3}{\circle*{4}}              
              \put(-5,23){$\delta(t-\red{t_K})$}
              \put(38,10){\framebox(40,16){$\red{a_K}$}}
              \put(0,18){\vector(1,0){37}}

              \put(78,103){\vector(1,-1){38}}
              \put(78,59){\vector(1,0){39}}
              \put(78,17){\vector(1,1){38}}

              \put(130,60){\circle{24}}
              \put(126,57){$\Large+$}
              \put(144,60){\vector(1,0){10}}
              \put(154,52){\framebox(50,16){$\Nc(t|0,v)$}}
              
              \put(205,75){$g_{\rthetab}(t)$}
              \put(204,60){\vector(1,0){33}}
              
              \put(252,85){$\epsilon(t)$}
              \put(250,95){\vector(0,-1){20}}
              \put(251,60){\circle{24}}
              \put(247,58){$\Large+$}
              
              \put(265,66){$g(t)$}
              \put(265,60){\vector(1,0){20}}
\end{picture}
}

\def\modeld{\tt\setlength{\unitlength}{0.92pt}
\begin{picture}(250,106)
\thinlines    \put(20,70){$s(t)=\sum_k \rak \delta(t-\red{t_k})$}
              \put(105,60){\vector(1,0){50}}
              \put(154,52){\framebox(50,16){$\Nc(t|0,v)$}}
              
              \put(205,75){$g_{\rthetab}(t)$}
              \put(204,60){\vector(1,0){33}}
              
              \put(252,85){$\epsilon(t)$}
              \put(250,95){\vector(0,-1){20}}
              \put(251,60){\circle{24}}
              \put(247,58){$\Large+$}
              
              \put(265,66){$g(t)$}
              \put(265,60){\vector(1,0){20}}
              \put(0,20){\siga}
\end{picture}
}

\def\siga{\setlength{\unitlength}{0.46pt}
\begin{picture}(290,108)
\thinlines    \put(10,26){\vector(1,0){260}}
              \put(16,12){\vector(0,1){88}}
              \put(48,27){\vector(0,1){34}}
              \put(79,27){\vector(0,1){54}}
              \put(130,27){\vector(0,1){46}}
              \put(188,26){\vector(0,1){40}}
              \put(262,14){$t$}
              \put(182,16){$\red{t_K}$}
              \put(195,63){$\red{a_K}$}
              \put(125,15){$\red{t_k}$}
              \put(137,68){$\red{a_k}$}
              \put(75,14){$\red{t_2}$}
              \put(84,73){$\red{a_2}$}
              \put(43,15){$\red{t_1}$}
              \put(54,56){$\red{a_1}$}
              \multiput(150,46)(8,0){3}{\circle*{4}}
\end{picture}
}

\def\modele{\tt\setlength{\unitlength}{0.92pt}
\begin{picture}(250,106)
\thinlines    \put(20,70){$s(t)=\sum_n \rfn \delta(t-n\Delta t)$}
              \put(105,60){\vector(1,0){50}}
              \put(154,52){\framebox(50,16){$\Nc(t|0,v)$}}
              
              \put(205,75){$g_{\rfb}(t)$}
              \put(204,60){\vector(1,0){33}}
              
              \put(252,85){$\epsilon(t)$}
              \put(250,95){\vector(0,-1){20}}
              \put(251,60){\circle{24}}
              \put(247,58){$\Large+$}
              
              \put(265,66){$g(t)$}
              \put(265,60){\vector(1,0){20}}
              \put(0,20){\sigb}
\end{picture}
}

\def\sigb{\setlength{\unitlength}{0.46pt}
\begin{picture}(290,108)
\thinlines    \put(10,26){\vector(1,0){260}}
              \put(10,12){\vector(0,1){88}}
              
              \put(20,27){\vector(0,1){34}}
              \put(40,27){\vector(0,1){14}}
              \put(60,27){\vector(0,1){54}}
              \put(80,27){\vector(0,1){31}}
              \put(100,27){\vector(0,1){46}}
              \put(120,27){\vector(0,1){40}}
              \put(140,27){\vector(0,1){48}}
              \put(160,27){\vector(0,1){34}}
              \put(180,27){\vector(0,1){14}}
              \put(200,27){\vector(0,1){31}}
              \put(220,27){\vector(0,1){46}}
              \put(240,27){\vector(0,1){40}}
              
              \put(270,14){$t$}
              \put(15,10){$1$}
              \put(15,63){$\red{f_1}$}
              \put(135,10){$n$}
              \put(135,68){$\red{f_n}$}
              \put(235,10){$N$}
              \put(235,68){$\red{f_N}$}

\end{picture}
}

\def\sigbr{\setlength{\unitlength}{0.46pt}
\begin{picture}(290,108)
\thinlines    \put(10,26){\vector(1,0){260}}
              \put(10,12){\vector(0,1){88}}
              
              \put(20,27){\vector(0,1){34}}
              \put(40,27){\vector(0,1){14}}
              \put(60,27){\vector(0,1){54}}
              \put(80,27){\vector(0,1){31}}
              \put(100,27){\vector(0,1){46}}
              \put(120,27){\vector(0,1){40}}
              \put(140,27){\vector(0,1){48}}
              \put(160,27){\vector(0,1){34}}
              \put(180,27){\vector(0,1){14}}
              \put(200,27){\vector(0,1){31}}
              \put(220,27){\vector(0,1){46}}
              \put(240,27){\vector(0,1){40}}
              \put(270,14){$t$}
              \put(15,10){$1$}
              \put(15,63){$\red{f_1}$}
              \put(135,10){$n$}
              \put(135,68){$\red{f_n}$}
              \put(235,10){$N$}
              \put(235,68){$\red{f_N}$}

              \put(60,10){$\red{\Delta t}$}
\end{picture}
}

\def\modelf{\tt\setlength{\unitlength}{0.92pt}
\begin{picture}(250,106)
\thinlines    \put(20,70){$s(t)=\sum_n \rfn \delta(t-n\red{\Delta t})$}
              \put(105,60){\vector(1,0){50}}
              \put(154,52){\framebox(50,16){$\Nc(t|0,\red{v})$}}
              
              \put(205,75){$g_{\rfb,\rthetab}(t)$}
              \put(204,60){\vector(1,0){33}}
              
              \put(252,85){$\epsilon(t)$}
              \put(250,95){\vector(0,-1){20}}
              \put(251,60){\circle{24}}
              \put(247,58){$\Large+$}
              
              \put(265,66){$g(t)$}
              \put(265,60){\vector(1,0){20}}
              \put(0,20){\sigbr}
\end{picture}
}

\def\sig{\tt\setlength{\unitlength}{1.38pt}
\begin{picture}(290,108)
\thinlines    
              \put(10,26){\vector(1,0){260}}
              \put(48,27){\vector(0,1){34}}
              \put(16,12){\vector(0,1){88}}
              \put(79,27){\vector(0,1){54}}
              \put(102,27){\vector(0,1){31}}
              \put(130,27){\vector(0,1){46}}
              \put(188,26){\vector(0,1){40}}
              \put(262,14){$t$}
              \put(22,90){$g(t)$}
              \put(182,16){$t_K$}
              \put(195,63){$a_K$}
              \put(125,15){$t_k$}
              \put(137,68){$a_k$}
              \put(75,14){$t_2$}
              \put(84,73){$a_2$}
              \put(43,15){$t_1$}
              \put(54,56){$a_1$}
              \multiput(150,46)(8,0){3}{\circle*{4}}
\end{picture}
}

\def\decgb{\tt\setlength{\unitlength}{0.92pt}
\begin{picture}(363,122)
\thinlines    \put(248,66){\vector(1,0){14}}
              \put(215,57){\framebox(33,18){$h(t)$}}
              \put(288,66){\vector(1,0){22}}
              \put(270,64){$\large\Sigma$}
              \put(275,67){\circle{26}}
              \put(275,102){\vector(0,-1){20}}
              \put(280,94){$b(t)$}
              \put(296,75){$y(t)$}
              \multiput(100,35)(0,8){3}{\circle*{4}}
              \put(194,72){$x(t)$}
              \put(193,66){\vector(1,0){23}}
              \put(176,61){$\large\Sigma$}
              \put(181,65){\circle{26}}
              \put(80,92){\framebox(43,18){$a_1$}}
              \put(80,58){\framebox(43,18){$a_2$}}
              \put(79,10){\framebox(43,18){$a_K$}}
              \put(123,66){\vector(1,0){45}}
              \put(123,101){\vector(4,-3){44}}
              \put(122,18){\vector(1,1){44}}
              \put(55,101){\vector(1,0){23}}
              \put(55,66){\vector(1,0){23}}
              \put(55,19){\vector(1,0){23}}
              \put(10,104){$g_1(t-t_1)$}
              \put(10,69){$g_2(t-t_2)$}
              \put(10,22){$g_K(t-t_k)$}
\end{picture}
}

\def\decgc{\tt\setlength{\unitlength}{0.92pt}
\begin{picture}(287,68)
\thinlines    \put(36,20){\vector(1,0){23}}
              \put(101,20){\vector(1,0){23}}
              \put(167,20){\vector(1,0){16}}
              \put(124,12){\framebox(43,18){$h(t)$}}
              \put(210,20){\vector(1,0){23}}
              \put(192,20){$\large\Sigma$}
              \put(196,23){\circle{26}}
              \put(196,57){\vector(0,-1){20}}
              \put(200,49){$b(t)$}
              \put(218,30){$y(t)$}
              \put(15,20){$s(t)$}
              \put(101,32){$x(t)$}
              \put(59,11){\framebox(43,18){$g(t)$}}
\end{picture}
}

\def\sigd{
\begin{picture}(290,108)
\thinlines    \put(10,26){\vector(1,0){260}}
              \put(48,27){\vector(0,1){34}}
              \put(16,12){\vector(0,1){88}}
              \put(79,27){\vector(0,1){54}}
              \put(102,27){\vector(0,1){31}}
              \put(130,27){\vector(0,1){46}}
              \put(188,26){\vector(0,1){40}}
              \put(262,14){$t$}
              \put(22,90){$g(t)$}
              \put(182,16){$t_K$}
              \put(195,63){$a_K$}
              \put(125,15){$t_k$}
              \put(137,68){$a_k$}
              \put(75,14){$t_2$}
              \put(84,73){$a_2$}
              \put(43,15){$t_1$}
              \put(54,56){$a_1$}
              \multiput(150,46)(8,0){3}{\circle*{4}}
\end{picture}
}

\def\wt#1{\widetilde{#1}}
\def\Thetab{\bm{\Theta}}
\def\Thetabh{\widehat{\Thetab}}
\def\abh{\widehat{\ab}}
\def\kbt{\tilde{\kb}}
\def\kt{\tilde{k}}
\def\gammat{\tilde{\gamma}}
\def\Sigmat{\tilde{\Sigma}}
\def\etat{\tilde{\eta}}

\def\rHb{\red{\Hb}}
\def\rFb{\red{\Fb}}

\def\wt#1{\widetilde{#1}}
\def\Thetab{\bm{\Theta}}
\def\Thetabh{\widehat{\Thetab}}
\def\abh{\widehat{\ab}}
\def\kbt{\tilde{\kb}}
\def\kt{\tilde{k}}
\def\gammat{\tilde{\gamma}}
\def\Sigmat{\tilde{\Sigma}}
\def\etat{\tilde{\eta}}
\def\zetat{\tilde{\zeta}}
\def\Mt{\tilde{M}}
\def\Mh{\hat{M}}
\def\Nt{\tilde{N}}
\def\atk{{\tilde{a}_k}}
\def\ztk{{\tilde{z}_k}}

\def\rh{\red{h}}
\def\rvbe{\rvb_{\epsilon}}
\def\rvei{\rv_{\epsilon_i}}
\def\rVbe{\rVb_{\epsilon}}
\def\rVbeh{\red{\Vbh}_{\epsilon}}
\def\vbh{\hat{\vb}}
\def\rVbfh{\red{\Vbh}_f}
\def\rvbeh{\rvbh_{\epsilon}}
\def\rvbet{\rvbt_{\epsilon}}
\def\rvbfh{\rvbh_f}

\def\rvbf{\rvb_f}
\def\rvfj{\rv_{f_j}}
\def\rVbf{\red{\Vb}_f}

\def\alphaez{\alpha_{\epsilon_0}}
\def\betaez{\beta_{\epsilon_0}}
\def\alphafz{\alpha_{f_0}}
\def\betafz{\beta_{f_0}}

\def\wh#1{\widehat{#1}}
\def\disp#1{{\displaystyle #1}}

\def\ER{\mbox{I\kern-.25em R}}
\def\EC{\mbox{C\kern-.8em C}}
\def\EZ{\mbox{Z\kern-.55em Z}}
\def\EN{\mbox{N\kern-.8em N}}

\def\bm#1{\mbox{\boldmath $#1$}}
\def\sb{{\bm s}}
\def\ub{{\bm u}}
\def\xb{{\bm x}}
\def\yb{{\bm y}}
\def\db{{\bm{d}}}
\def\Ab{{\bm A}}
\def\Bb{{\bm B}}
\def\Fb{{\bm F}}
\def\Rb{{\bm R}}
\def\Sb{{\bm S}}
\def\Ub{{\bm U}}

\def\thetab{\bm{\theta}}
\def\phib{\bm{\phi}}
\def\lambdab{\bm{\lambda}}
\def\Lambdab{\bm{\Lambda}}
\def\Sigmab{\bm{\Sigma}}
\def\lra{\longrightarrow}
\def\intg{\int\kern-1.1em\int}
\def\intd{\int\kern-.8em\int}

\def\apriori{{\em a priori} }
\def\aposteriori{{\em a posteriori} }

\def\d#1{\,\mbox{d}#1}
\def\esp#1{\mbox{E}\left\{ #1 \right\}}
\def\espx#1#2{\mbox{E}_{#1}\left\{ #2 \right\}}
\def\expf#1{\exp\left[ {#1} \right]}
\def\dpdx#1#2{\frac{\partial #1}{\partial #2}}
\def\dpdxd#1#2{\frac{\partial^2 #1}{\partial #2^2}}

\def\ent#1{\hbox{H}\left[#1\right]}
\def\dent#1{\delta\hbox{H}\left[#1\right]}
\def\rent#1{\hbox{D}\left[#1\right]}
\def\kullback#1{\hbox{K}\left[#1\right]}
\def\pent#1{\hbox{PE}\left[#1\right]}
\def\mdiv#1{\hbox{J}\left[#1\right]}
\def\infmut#1{\hbox{I}\left[#1\right]}
\def\idisc#1{\hbox{ID}\left[#1\right]}
\def\j{\mbox{j}}

\def\pxtheta{p(x | \thetab)}
\def\pxthetas{p(x | \thetab^*)}
\def\pxthetasd{p(x | \thetab^*+\Delta\thetab)}

\def\tr#1{\hbox{tr}\left(#1\right)}
\def\det#1{\hbox{det}\left(#1\right)}

\def\ieeeSSC{IEEE Transactions on Systems Science and Cybernetics}
\def\ieeeSP{IEEE Transactions on Signal Processing}			
\def\ieeeP{Proceedings of the IEEE}					
\def\ieeeIT{IEEE Transactions on Information Theory}			
\def\TS{Traitement du Signal}						
\def\siamCO{SIAM Journal of Control}					
\def\AISM{Annals of Institute of Statistical Mathematics}		
\def\ieeeASSP{IEEE Transactions on Acoustics Speech and Signal Processing}
\def\ieeeAC{IEEE Transactions on Automatic and Control}			

\def\jan{janvier\xspace}
\def\feb{f\'evrier\xspace}
\def\mar{mars\xspace}
\def\apr{avril\xspace}
\def\may{mai\xspace}
\def\jun{juin\xspace}
\def\jul{juillet\xspace}
\def\aug{ao\^ut\xspace}
\def\sep{septembre\xspace}
\def\oct{octobre\xspace}
\def\nov{novembre\xspace}
\def\dec{d\'ecembre\xspace}
\def\Jan{January\xspace}	
\def\Feb{February\xspace}
\def\Mar{March\xspace}
\def\Apr{April\xspace}
\def\May{May\xspace}
\def\Jun{June\xspace}
\def\Jul{July\xspace}
\def\Aug{August\xspace}
\def\Sep{September\xspace}
\def\Oct{October\xspace}
\def\Nov{November\xspace}
\def\Dec{December\xspace}

\def\espq#1{\left<#1\right>_q}

\def\rZbe{\rZb_{\epsilon}}
\def\nubt{\widetilde{\nub}}

\def\rVbf{\rVb_{f}}
\def\rvbf{\rvb_{f}}
\def\rvfj{\red{v}_{f_j}}
\def\rvbhe{\rvbh_{\epsilon}}
\def\rvbhf{\rvbh_{f}}
\def\bsplit{\begin{split}}
\def\esplit{\end{split}}
\def\alphaei{\alpha_{\epsilon_i}}
\def\alphafj{\alpha_{f_j}}
\def\betaei{\beta_{\epsilon_i}}
\def\betafj{\beta_{f_j}}

\def\bgbh{\widehat{\bgb}}
\def\rfh{\widehat{\rf}}
\def\rFh{\widehat{\rF}}
\def\rc{\red{r}}
\def\rch{\widehat{\rc}}
\def\partialx{\frac{\partial }{\partial x}}
\def\partialy{\frac{\partial }{\partial y}}
\def\partialr{\frac{\partial }{\partial r}}
\def\partialxd{\frac{\partial^2 }{\partial x^2}}
\def\partialyd{\frac{\partial^2 }{\partial y^2}}
\def\partialrd{\frac{\partial^2 }{\partial r^2}}
\def\cosphi{\cos(\phi)}
\def\sinphi{\sin(\phi)}
\def\sumk{\sum_{k=1}^K}
\def\sumj{\sum_{j=1}^N}
\def\sumi{\sum_{i=1}^M}
\def\summ{\sum_{m=1}^M}
\def\sumn{\sum_{n=1}^N}
\def\summn{\sum_{m=1}^M\sum_{n=1}^N}

\def\df{\dot{f}}
\def\ddf{\ddot{f}}

\def\rf{\red{f}}
\def\rdf{\red{\df}}
\def\rddf{\red{\ddf}}

\def\rfb{\red{\fb}}
\def\rdfb{\red{\dot{\fb}}}
\def\rddfb{\red{\ddot{\fb}}}

\def\bdg{\blue{\dot{g}}}
\def\bddg{\blue{\ddot{g}}}

\def\bgb{\blue{\gb}}
\def\bdg{\blue{\dot{g}}}

\def\bdgb{\blue{\dot{\gb}}}
\def\bddgb{\blue{\ddot{\gb}}}

\def\epsilonbd{\dot{\epsilonb}}
\def\epsilonbdd{\ddot{\epsilonb}}

\def\rdfbh{\widehat{\rdfb}}
\def\rddfbh{\widehat{\rddfb}}

\def\rqh{\red{\widehat{q}}}
\def\rdfh{\red{\widehat{\dot{f}}}}
\def\rddfh{\red{\widehat{\ddot{f}}}}

\def\vek{v_{\epsilon_k}}
\def\rveih{\widehat{\rve}_i}
\def\rvfjh{\widehat{\rvf}_j}

\def\alphaxiz{{\alpha_{\xi_0}}}
\def\betaxiz{{\beta_{\xi_0}}}
\def\alphazz{{\alpha_{z_0}}}

\def\betazz{{\beta_{z_0}}}
\def\rVbz{{\rVb_z}}
\def\rvbz{{\rvb_z}}
\def\rvxi{{\rv_\xi}}
\def\rvzj{{\rv_{z_j}}}
\def\rvxih{{\widehat{\rv}_\xi}}
\def\rvbzh{{\widehat{\rvb}_z}}
\def\rvzjh{{\widehat{\rv}_{z_j}}}

\def\alphae{{\alpha_\epsilon}}

\def\alphaet{{\widetilde{\alpha}_\epsilon}}
\def\alphaft{{\widetilde{\alpha}_f}}
\def\betae{{\beta_\epsilon}}
\def\betaet{{\widetilde{\beta}_\epsilon}}
\def\betaft{{\widetilde{\beta}_f}}
\def\rfih{\hat{\rf}_i}
\def\rfbk{\rfb^{(k)}}
\def\rfbkp{\rfb^{(k+1)}}

\def\rmh{\hat{\red{m}}}

\def\rmk{{\red{m}_k}}
\def\rmkh{{\hat{\red{m}}_k}}
\def\rvk{{\red{v}_k}}
\def\rvkh{{\hat{\red{v}}_k}}
\def\rak{{\red{a}_k}}
\def\rakh{{\hat{\red{a}}_k}}

\def\rvh{\hat{\red{v}}}
\def\rveh{\hat{\rve}}

\def\rmt{\tilde{\red{m}}}
\def\rvt{\tilde{\red{v}}}
\def\rvet{\tilde{\rve}}
\def\rveih{\red{\hat{v}}_{\epsilon_i}}

\def\rveb{\red{\vb_{\epsilon}}}
\def\rvebh{\red{\vbh_{\epsilon}}}

\def\blocgi{\begin{picture}(80,50)
  \put(60,40){\makebox(0,0){$\red{m}$}}
  \put(60,40){\circle{20}}
  \put(60,30){\vector(0,-1){20}}
  \put(60,0){\circle{20}}
  \put(60,0){\makebox(0,0){$\blue{g}_i$}}
  \put(20,0){\circle{20}}
  \put(20,0){\makebox(0,0){$\epsilon_i$}}
  \put(30,0){\vector(1,0){20}}
\end{picture}
}
\def\blocgfi{\begin{picture}(80,50)
  \put(60,40){\makebox(0,0){$\rf_i$}}
  \put(60,40){\circle{20}}
  \put(60,30){\vector(0,-1){20}}
  \put(60,0){\circle{20}}
  \put(60,0){\makebox(0,0){$\blue{g}_i$}}
  \put(20,0){\circle{20}}
  \put(20,0){\makebox(0,0){$\epsilon_i$}}
  \put(30,0){\vector(1,0){20}}
\end{picture}
}
\def\blocghfi{\begin{picture}(80,50)
  \put(60,40){\makebox(0,0){$\rf_i$}}
  \put(60,40){\circle{20}}
  \put(60,30){\vector(0,-1){20}}
  \put(65,20){\makebox(0,0){$h$}}
  \put(60,0){\circle{20}}
  \put(60,0){\makebox(0,0){$\blue{g}_i$}}
  \put(20,0){\circle{20}}
  \put(20,0){\makebox(0,0){$\epsilon_i$}}
  \put(30,0){\vector(1,0){20}}
\end{picture}
}
\def\blocgie{\begin{picture}(80,50)
  \put(60,40){\makebox(0,0){$\red{m}$}}
  \put(60,40){\circle{20}}
  \put(60,30){\vector(0,-1){20}}
  \put(60,0){\circle{20}}
  \put(60,0){\makebox(0,0){$\blue{g}_i$}}
  \put(30,0){\circle{20}}
  \put(30,0){\makebox(0,0){$\epsilon_i$}}
  \put(40,0){\vector(1,0){10}}
  \put(0,0){\circle{20}}
  \put(0,0){\makebox(0,0){$\rve$}}
  \put(10,0){\vector(1,0){10}}
\end{picture}
}
\def\blocgievei{\begin{picture}(80,50)
  \put(60,40){\makebox(0,0){$\red{m}$}}
  \put(60,40){\circle{20}}
  \put(60,30){\vector(0,-1){20}}
  \put(60,0){\circle{20}}
  \put(60,0){\makebox(0,0){$\blue{g}_i$}}
  \put(30,0){\circle{20}}
  \put(30,0){\makebox(0,0){$\epsilon_i$}}
  \put(40,0){\vector(1,0){10}}
  \put(0,0){\circle{20}}
  \put(0,0){\makebox(0,0){$\rvei$}}
  \put(10,0){\vector(1,0){10}}
\end{picture}
}
\def\blocfgie{\begin{picture}(80,100)
  \put(60,40){\makebox(0,0){$\rf_i$}}
  \put(60,40){\circle{20}}
  \put(60,30){\vector(0,-1){20}}
  \put(65,20){\makebox(0,0){$\blue{h}$}}
  
  \put(60,0){\circle{20}}
  \put(60,0){\makebox(0,0){$\blue{g}_i$}}
  \put(30,0){\circle{20}}
  \put(30,0){\makebox(0,0){$\epsilon_i$}}
  \put(40,0){\vector(1,0){10}}
  \put(0,0){\circle{20}}
  \put(0,0){\makebox(0,0){$\rve$}}
  \put(10,0){\vector(1,0){10}}
\end{picture}
}
\def\blocvfgie{\begin{picture}(80,100)
  \put(60,80){\makebox(0,0){$\rv_i$}}
  \put(60,80){\circle{20}}
  \put(60,70){\vector(0,-1){20}}
  
  \put(60,40){\makebox(0,0){$\rf_i$}}
  \put(60,40){\circle{20}}
  \put(60,30){\vector(0,-1){20}}
  \put(65,20){\makebox(0,0){$\blue{h}$}}
  
  \put(60,0){\circle{20}}
  \put(60,0){\makebox(0,0){$\blue{g}_i$}}
  \put(30,0){\circle{20}}
  \put(30,0){\makebox(0,0){$\epsilon_i$}}
  \put(40,0){\vector(1,0){10}}
  \put(0,0){\circle{20}}
  \put(0,0){\makebox(0,0){$\rve$}}
  \put(10,0){\vector(1,0){10}}
\end{picture}
}
\def\blocvfhgie{\begin{picture}(80,100)
  \put(60,80){\makebox(0,0){$\rv_i$}}
  \put(60,80){\circle{20}}
  \put(60,70){\vector(0,-1){20}}
  
  \put(60,40){\makebox(0,0){$\rf_i$}}
  \put(60,40){\circle{20}}
  \put(60,30){\vector(0,-1){20}}
  
  \put(30,30){\circle{20}}
  \put(30,30){\makebox(0,0){$\red{h}$}}
  \put(38,22){\vector(1,-1){15}}
  
  \put(60,0){\circle{20}}
  \put(60,0){\makebox(0,0){$\blue{g}_i$}}
  \put(30,0){\circle{20}}
  \put(30,0){\makebox(0,0){$\epsilon_i$}}
  \put(40,0){\vector(1,0){10}}
  \put(0,0){\circle{20}}
  \put(0,0){\makebox(0,0){$\rve$}}
  \put(10,0){\vector(1,0){10}}
\end{picture}
}
\def\blocvfhgiei{\begin{picture}(80,100)
  \put(60,80){\makebox(0,0){$\rv_i$}}
  \put(60,80){\circle{20}}
  \put(60,70){\vector(0,-1){20}}
  
  \put(60,40){\makebox(0,0){$\rf_i$}}
  \put(60,40){\circle{20}}
  \put(60,30){\vector(0,-1){20}}
  
  \put(30,30){\circle{20}}
  \put(30,30){\makebox(0,0){$\red{h}$}}
  \put(38,22){\vector(1,-1){15}}
  
  \put(60,0){\circle{20}}
  \put(60,0){\makebox(0,0){$\blue{g}_i$}}
  \put(30,0){\circle{20}}
  \put(30,0){\makebox(0,0){$\epsilon_i$}}
  \put(40,0){\vector(1,0){10}}
  \put(0,0){\circle{20}}
  \put(0,0){\makebox(0,0){$\rvei$}}
  \put(10,0){\vector(1,0){10}}
\end{picture}
}
\def\blocfgHe{\begin{picture}(80,100)
  \put(60,40){\makebox(0,0){$\rfb$}}
  \put(60,40){\circle{20}}
  \put(60,30){\vector(0,-1){20}}
  \put(65,20){\makebox(0,0){$\blue{\Hb}$}}
  
  \put(60,0){\circle{20}}
  \put(60,0){\makebox(0,0){$\bgb$}}
  \put(30,0){\circle{20}}
  \put(30,0){\makebox(0,0){$\epsilonb$}}
  \put(40,0){\vector(1,0){10}}
  \put(0,0){\circle{20}}
  \put(0,0){\makebox(0,0){$\rve$}}
  \put(10,0){\vector(1,0){10}}
\end{picture}
}
\def\blocvfgHe{\begin{picture}(80,100)
  \put(60,80){\makebox(0,0){$\rvb$}}
  \put(60,80){\circle{20}}
  \put(60,70){\vector(0,-1){20}}
  
  \put(60,40){\makebox(0,0){$\rfb$}}
  \put(60,40){\circle{20}}
  \put(60,30){\vector(0,-1){20}}
  \put(65,20){\makebox(0,0){$\bHb$}}
  
  \put(60,0){\circle{20}}
  \put(60,0){\makebox(0,0){$\bgb$}}
  \put(30,0){\circle{20}}
  \put(30,0){\makebox(0,0){$\epsilonb$}}
  \put(40,0){\vector(1,0){10}}
  \put(0,0){\circle{20}}
  \put(0,0){\makebox(0,0){$\rve$}}
  \put(10,0){\vector(1,0){10}}
\end{picture}
}
\def\blocvfgHei{\begin{picture}(80,100)
  \put(60,80){\makebox(0,0){$\rvb$}}
  \put(60,80){\circle{20}}
  \put(60,70){\vector(0,-1){20}}
  
  \put(60,40){\makebox(0,0){$\rfb$}}
  \put(60,40){\circle{20}}
  \put(60,30){\vector(0,-1){20}}
  \put(65,20){\makebox(0,0){$\bHb$}}
  
  \put(60,0){\circle{20}}
  \put(60,0){\makebox(0,0){$\bgb$}}
  \put(30,0){\circle{20}}
  \put(30,0){\makebox(0,0){$\epsilonb$}}
  \put(40,0){\vector(1,0){10}}
  \put(0,0){\circle{20}}
  \put(0,0){\makebox(0,0){$\rvei$}}
  \put(10,0){\vector(1,0){10}}
\end{picture}
}
\def\blocvfHge{\begin{picture}(80,100)
  \put(60,80){\makebox(0,0){$\rvb$}}
  \put(60,80){\circle{20}}
  \put(60,70){\vector(0,-1){20}}
  
  \put(60,40){\makebox(0,0){$\rfb$}}
  \put(60,40){\circle{20}}
  \put(60,30){\vector(0,-1){20}}
  
  \put(30,30){\circle{20}}
  \put(30,30){\makebox(0,0){$\rHb$}}
  \put(38,22){\vector(1,-1){15}}
  
  \put(60,0){\circle{20}}
  \put(60,0){\makebox(0,0){$\bgb$}}
  \put(30,0){\circle{20}}
  \put(30,0){\makebox(0,0){$\epsilonb$}}
  \put(40,0){\vector(1,0){10}}
  \put(0,0){\circle{20}}
  \put(0,0){\makebox(0,0){$\rve$}}
  \put(10,0){\vector(1,0){10}}
\end{picture}
}
\def\blocvfHgei{\begin{picture}(80,100)
  \put(60,80){\makebox(0,0){$\rvb$}}
  \put(60,80){\circle{20}}
  \put(60,70){\vector(0,-1){20}}
  
  \put(60,40){\makebox(0,0){$\rfb$}}
  \put(60,40){\circle{20}}
  \put(60,30){\vector(0,-1){20}}
  
  \put(30,30){\circle{20}}
  \put(30,30){\makebox(0,0){$\rHb$}}
  \put(38,22){\vector(1,-1){15}}
  
  \put(60,0){\circle{20}}
  \put(60,0){\makebox(0,0){$\bgb$}}
  \put(30,0){\circle{20}}
  \put(30,0){\makebox(0,0){$\epsilonb$}}
  \put(40,0){\vector(1,0){10}}
  \put(0,0){\circle{20}}
  \put(0,0){\makebox(0,0){$\rvei$}}
  \put(10,0){\vector(1,0){10}}
\end{picture}
}
\def\blocvhFge{\begin{picture}(80,100)
  \put(60,80){\makebox(0,0){$\rvb$}}
  \put(60,80){\circle{20}}
  \put(60,70){\vector(0,-1){20}}
  
  \put(60,40){\makebox(0,0){$\rhb$}}
  \put(60,40){\circle{20}}
  \put(60,30){\vector(0,-1){20}}
  
  \put(30,30){\circle{20}}
  \put(30,30){\makebox(0,0){$\bFb$}}
  \put(38,22){\vector(1,-1){15}}

  \put(60,0){\circle{20}}
  \put(60,0){\makebox(0,0){$\bgb$}}
  \put(30,0){\circle{20}}
  \put(30,0){\makebox(0,0){$\epsilonb$}}
  \put(40,0){\vector(1,0){10}}
  \put(0,0){\circle{20}}
  \put(0,0){\makebox(0,0){$\rve$}}
  \put(10,0){\vector(1,0){10}}
\end{picture}
}
\def\blocvhFgei{\begin{picture}(80,100)
  \put(60,80){\makebox(0,0){$\rvb$}}
  \put(60,80){\circle{20}}
  \put(60,70){\vector(0,-1){20}}
  
  \put(60,40){\makebox(0,0){$\rhb$}}
  \put(60,40){\circle{20}}
  \put(60,30){\vector(0,-1){20}}
  
  \put(30,30){\circle{20}}
  \put(30,30){\makebox(0,0){$\bFb$}}
  \put(38,22){\vector(1,-1){15}}

  \put(60,0){\circle{20}}
  \put(60,0){\makebox(0,0){$\bgb$}}
  \put(30,0){\circle{20}}
  \put(30,0){\makebox(0,0){$\epsilonb$}}
  \put(40,0){\vector(1,0){10}}
  \put(0,0){\circle{20}}
  \put(0,0){\makebox(0,0){$\rvei$}}
  \put(10,0){\vector(1,0){10}}
\end{picture}
}
\def\blochFge{\begin{picture}(80,40)
  \put(60,40){\makebox(0,0){$\rhb$}}
  \put(60,40){\circle{20}}
  \put(60,30){\vector(0,-1){20}}
  
  \put(30,30){\circle{20}}
  \put(30,30){\makebox(0,0){$\bFb$}}
  \put(38,22){\vector(1,-1){15}}

  \put(60,0){\circle{20}}
  \put(60,0){\makebox(0,0){$\bgb$}}
  \put(30,0){\circle{20}}
  \put(30,0){\makebox(0,0){$\epsilonb$}}
  \put(40,0){\vector(1,0){10}}
  \put(0,0){\circle{20}}
  \put(0,0){\makebox(0,0){$\rve$}}
  \put(10,0){\vector(1,0){10}}
\end{picture}
}
\def\blochFgei{\begin{picture}(80,40)
  \put(60,40){\makebox(0,0){$\rhb$}}
  \put(60,40){\circle{20}}
  \put(60,30){\vector(0,-1){20}}
  
  \put(30,30){\circle{20}}
  \put(30,30){\makebox(0,0){$\bFb$}}
  \put(38,22){\vector(1,-1){15}}

  \put(60,0){\circle{20}}
  \put(60,0){\makebox(0,0){$\bgb$}}
  \put(30,0){\circle{20}}
  \put(30,0){\makebox(0,0){$\epsilonb$}}
  \put(40,0){\vector(1,0){10}}
  \put(0,0){\circle{20}}
  \put(0,0){\makebox(0,0){$\rvei$}}
  \put(10,0){\vector(1,0){10}}
\end{picture}
}
\def\blocfHge{\begin{picture}(80,40)  
  \put(60,40){\makebox(0,0){$\bfb$}}
  \put(60,40){\circle{20}}
  \put(60,30){\vector(0,-1){20}}
  
  \put(30,30){\circle{20}}
  \put(30,30){\makebox(0,0){$\rHb$}}
  \put(38,22){\vector(1,-1){15}}
  
  \put(60,0){\circle{20}}
  \put(60,0){\makebox(0,0){$\bgb$}}
  \put(30,0){\circle{20}}
  \put(30,0){\makebox(0,0){$\epsilonb$}}
  \put(40,0){\vector(1,0){10}}
  \put(0,0){\circle{20}}
  \put(0,0){\makebox(0,0){$\rve$}}
  \put(10,0){\vector(1,0){10}}
\end{picture}
}
\def\blocfHgei{\begin{picture}(80,40)  
  \put(60,40){\makebox(0,0){$\bfb$}}
  \put(60,40){\circle{20}}
  \put(60,30){\vector(0,-1){20}}
  
  \put(30,30){\circle{20}}
  \put(30,30){\makebox(0,0){$\rHb$}}
  \put(38,22){\vector(1,-1){15}}
  
  \put(60,0){\circle{20}}
  \put(60,0){\makebox(0,0){$\bgb$}}
  \put(30,0){\circle{20}}
  \put(30,0){\makebox(0,0){$\epsilonb$}}
  \put(40,0){\vector(1,0){10}}
  \put(0,0){\circle{20}}
  \put(0,0){\makebox(0,0){$\rvei$}}
  \put(10,0){\vector(1,0){10}}
\end{picture}
}
\def\blocFgHe{\begin{picture}(80,40)
  \put(60,40){\makebox(0,0){$\bfb$}}
  \put(60,40){\circle{20}}
  \put(60,30){\vector(0,-1){20}}
  \put(65,20){\makebox(0,0){$\blue{\rHb}$}}
  
  \put(60,0){\circle{20}}
  \put(60,0){\makebox(0,0){$\bgb$}}
  \put(30,0){\circle{20}}
  \put(30,0){\makebox(0,0){$\epsilonb$}}
  \put(40,0){\vector(1,0){10}}
  \put(0,0){\circle{20}}
  \put(0,0){\makebox(0,0){$\rve$}}
  \put(10,0){\vector(1,0){10}}
\end{picture}
}

\def\ftrfi{\bg(r,\phi)}
\def\wth{\widehat{\tilde{h}}}
\def\varx#1#2{\mbox{Var}_{#1}\left\{#2\right\}}
\def\rvbxi{\rvb_{\xi}}
\def\rVbxi{\rVb_{\xi}}
\def\rvbxih{\rvbh_{\xi}}

\def\repsilon{\red{\epsilon}}

\def\vzj{v_{z_j}}
\def\vxij{v_{\xi_j}}
\def\vei{v_{\epsilon_i}}

\def\vzjh{\widehat{v}_{z_j}}
\def\vxijh{\widehat{v}_{\xi_j}}
\def\veih{\widehat{v}_{\epsilon_i}}

\def\vbeh{\widehat{\vb}_{\epsilon}}
\def\vbxih{\widehat{\vb}_{\xi}}
\def\vbzh{\widehat{\vb}_{z}}

\def\Vbeh{\widehat{\Vb}_{\epsilon}}
\def\Vbxih{\widehat{\Vb}_{\xi}}
\def\Vbzh{\widehat{\Vb}_{z}}

\def\Ybeh{\widehat{\Yb}_{\epsilon}}
\def\Ybxih{\widehat{\Yb}_{\xi}}
\def\Ybzh{\widehat{\Yb}_{z}}

\def\vf{v_f}
\def\vbe{\vb_{\epsilon}}
\def\vbz{\vb_z}
\def\vxi{v_{\xi}}
\def\vbxi{\vb_{\xi}}
\def\Vbe{\Vb_{\epsilon}}
\def\Vbxi{\Vb_{\xi}}
\def\Vbz{\Vb_z}

\def\half{\frac{1}{2}}

\def\dpdx#1#2{\frac{\partial {#1}}{\partial {#2}}}
\def\dpdxd#1#2{\frac{\partial^2 {#1}}{\partial {#2}^2}}
\def\dpdxdy#1#2#3{{{\partial ^2 {#1}}\over{\partial {#2} \partial {#3}}}}

\def\KL#1#2{\mbox{KL}\left[#1 : #2\right]}
\def\Covx#1#2{\mbox{Cov}_{#1}(#2)}
\def\braket#1#2{<{#1}(#2)>}

\def\rlambdab{\red{\lambdab}}
\def\bve{\blue{\ve}}
\def\ralphafjt{\red{\tilde{\alpha}_{f_j}}}
\def\rbetafjt{\red{\tilde{\beta}_{f_j}}}
\def\rSigmabt{\red{\tilde{\Sigmab}}}
\def\rfjt{\red{\tilde{f_j}}}
\def\zetab{\bm{\zeta}}
\def\bvxi{\blue{v_\xi}}
\def\bgamma{\blue{\gamma}}
\def\rgamma{\red{\gamma}}
\def\alphagamz{\alpha_{\gamma_0}}
\def\betagamz{\beta_{\gamma_0}}
\def\alphazetaz{\alpha_{\zeta_0}}
\def\betazetaz{\beta_{\zeta_0}}

\def\bvf{\blue{v_f}}
\def\bvzeta{\blue{v_\zeta}}
\def\rgb{\red{\gb}}
\def\rvxii{\red{v_{\xi_i}}}
\def\rvzeta{\red{v_{\zeta}}}

\usepackage{tikz,mathpazo}
\usetikzlibrary{shapes.geometric, arrows}
\usetikzlibrary{positioning,shapes,shadows,arrows}
\usetikzlibrary{mindmap,backgrounds}
\tikzstyle{startstop} = [rectangle, rounded corners, minimum width=1.5cm, minimum height=0.5cm,text centered, draw=black]
\tikzstyle{io} = [trapezium, trapezium left angle=70, trapezium right angle=110, minimum width=1.5cm, minimum height=0.6cm, text centered, draw=black]
\tikzstyle{process} = [rectangle, minimum width=1cm, minimum height=0.5cm, text centered, draw=black]
\tikzstyle{note} = [text centered]
\tikzstyle{decision} = [diamond, minimum width=1.5cm, minimum height=0.3cm, text centered, draw=black]
\tikzstyle{arrow} = [thick,->,>=stealth]

\def\blambda{\blue{\lambda}}
\def\blambdab{\blue{\lambdab}}

\def\fxy{f(x,y)}
\def\afxy{a\left(f(x,y)\right)}

\def\hxy{h(x,y)}
\def\gxy{g(x,y)}

\def\gt{\widetilde{g}}
\def\ft{\widetilde{f}}
\def\htxy{\widetilde{h}(x,y)}
\def\htzero{\widetilde{h}_0}
\def\htone{\widetilde{h}_1}
\def\httwo{\widetilde{h}_2}

\def\hdxy{h_d(x,y)}
\def\gdxy{g_d(x,y)}

\def\exy{\epsilon(x,y)}
\def\nxy{n(x,y)}
\def\rxy{r(x,y)}

\def\dxy{d(x,y)}
\def\dhxy{d_h(x,y)}
\def\dfxy{d_f(x,y)}

\def\fhxy{\hat{f}(x,y)}
\def\ghxy{\hat{f}(x,y)}
\def\ehxy{\hat{\epsilon}(x,y)}
\def\nhxy{\hat{n}(x,y)}

\def\ftuv{\widetilde{f}(u,v)}
\def\htuv{\widetilde{h}(u,v)}
\def\gtuv{\widetilde{g}(u,v)}
\def\rtuv{\widetilde{r}(u,v)}
\def\ntuv{\widetilde{n}(u,v)}
\def\dtuv{\widetilde{d}(u,v)}
\def\dhtuv{\widetilde{d}_h(u,v)}
\def\dftuv{\widetilde{d}_f(u,v)}

\def\htcuv{\widetilde{h}^*(u,v)}
\def\ftcuv{\widetilde{f}^*(u,v)}

\def\ftuv{\widetilde{f}(u,v)}
\def\htuv{\widetilde{h}(u,v)}
\def\etuv{\widetilde{\epsilon}(u,v)}

\def\ftcuv{\widetilde{f}^*(u,v)}
\def\htcuv{\widetilde{h}^*(u,v)}
\def\etcuv{\widetilde{\epsilon}^*(u,v)}

\def\fhtuv{\widehat{\widetilde{f}}(u,v)}
\def\hhtuv{\widehat{\widetilde{h}}(u,v)}

\def\fhtxy{\widehat{\widetilde{f}}(x,y)}
\def\hhtxy{\widehat{\widetilde{h}}(x,y)}

\def\fhtcuv{\widehat{\widetilde{f}}^*(u,v)}
\def\hhtcuv{\widehat{\widetilde{h}}^*(u,v)}
\def\ghtcuv{\widehat{\widetilde{g}}^*(u,v)}

\def\htzerouv{\widetilde{h}_0(u,v)}
\def\htoneuv{\widetilde{h}_1(u,v)}
\def\httwouv{\widetilde{h}_2(u,v)}
\def\htzerocuv{\widetilde{h}_0^*(u,v)}
\def\htonecuv{\widetilde{h}_1^*(u,v)}
\def\httwocuv{\widetilde{h}_2^*(u,v)}
\def\gtoneuv{\widetilde{g}_1(u,v)}
\def\gttwouv{\widetilde{g}_2(u,v)}

\def\ftuvk{\widetilde{f}^{(k)}(u,v)}
\def\htuvk{\widetilde{h}^{(k)}(u,v)}

\def\fhxyk{\widehat{f}^{(k)}(x,y)}
\def\hhxyk{\widehat{h}^{(k)}(x,y)}
\def\fhxykp{\widehat{f}^{(k+1)}(x,y)}
\def\hhxykp{\widehat{h}^{(k)}(x,y)}

\def\fhk{\widehat{f}^{(k)}}
\def\hhk{\widehat{h}^{(k)}}
\def\fhkp{\widehat{f}^{(k+1)}}
\def\hhkp{\widehat{h}^{(k)}}

\def\fxyk{f^{(k)}(x,y)}
\def\fxykp{f^{(k+1)}(x,y)}

\def\fkxy{f^{(k)}(x,y)}
\def\fkpxy{f^{(k+1)}(x,y)}

\def\ftuvkp{\widetilde{f}^{(k+1)}(u,v)}
\def\htuvkp{\widetilde{h}^{(k+1)}(u,v)}

\def\hktuv{\widetilde{h}_k(u,v)}
\def\gktuv{\widetilde{g}_k(u,v)}

\def\bfxy{\blue{f}(x,y)}
\def\bhxy{\blue{h}(x,y)}
\def\bgxy{\blue{g}(x,y)}

\def\rfxy{\red{f}(x,y)}
\def\rhxy{\red{h}(x,y)}

\def\rftuv{\widetilde{\red{f}}(u,v)}
\def\rhtuv{\widetilde{\red{h}}(u,v)}

\def\bgtuv{\widetilde{\blue{g}}(u,v)}

\def\alphabh{\widehat{\alphab}}

\def\dfdx#1#2{\frac{\partial #1}{\partial #2}}

\def\zxy{z(x,y)}
\def\Txy{T(x,y)}
\def\fxy{f(x,y)}
\def\gzxy{g_0(x,y)}
\def\gxy{g(x,y)}
\def\phixy{\phi(x,y)}
\def\exy{e(x,y)}
\def\epsxy{\epsilon(x,y)}

\def\zhxy{\hat{z}(x,y)}
\def\Thxy{\hat{T}(x,y)}
\def\gzhxy{\hat{g}_0(x,y)}
\def\fhxy{\hat{f}(x,y)}
\def\phihxy{\hat{\phi}(x,y)}
\def\epshxy{\hat{\epsilon}(x,y)}